\documentclass[runningheads]{llncs}

\usepackage{eccv}

\usepackage{eccvabbrv}

\usepackage{graphicx}
\usepackage{booktabs}
\usepackage[inkscapelatex=false]{svg}

\usepackage[accsupp]{axessibility}  

\usepackage[pagebackref,breaklinks,colorlinks,citecolor=eccvblue]{hyperref}

\usepackage{orcidlink}
\usepackage{multirow}
\usepackage{xcolor} 
\usepackage[marginal]{footmisc}

\definecolor{darkgreen}{rgb}{0.0, 0.5, 0.0} 
\definecolor{darkgray}{rgb}{0.4, 0.4, 0.4}
\definecolor{darkred}{RGB}{153, 0, 0}

\usepackage[dvipsnames]{xcolor}

\begin{document}

\title{FoodLMM: A Versatile Food Assistant using Large Multi-modal Model} 


\author{
Yuehao Yin\textsuperscript{\rm 1$\dagger$}, 
Huiyan Qi\textsuperscript{\rm 1$\dagger$}, 
Bin Zhu\textsuperscript{\rm 2}, 
Jingjing Chen\textsuperscript{\rm 1*}, 
Yu-Gang Jiang\textsuperscript{\rm 1}, 
Chong-Wah Ngo\textsuperscript{\rm 2}\\
}

\authorrunning{Y Yin, H Qi, et al.}

\institute{School of Computer Science, Fudan University \and
Singapore Management University\\
\email{\{yhyin21, huiyanqi21\}@m.fudan.edu.cn, \{chenjingjing, ygj\}@fudan.edu.cn, \{binzhu, cwngo\}@smu.edu.sg}
}

\maketitle
\vspace{-0.3cm}
\begin{abstract}
Large Multi-modal Models (LMMs) have made impressive progress in many vision-language tasks. Nevertheless, the performance of general LMMs in specific domains is still far from satisfactory. This paper proposes FoodLMM, a versatile food assistant based on LMMs with various capabilities, including food recognition, ingredient recognition, recipe generation, nutrition estimation, food segmentation and multi-round conversation. To facilitate FoodLMM to deal with tasks beyond pure text output, we introduce a series of novel task-specific tokens and heads, enabling the model to predict food nutritional values and multiple segmentation masks. We adopt a two-stage training strategy. In the first stage, we utilize multiple public food benchmarks for multi-task learning by leveraging the instruct-following paradigm. In the second stage, we construct a multi-round conversation dataset and a reasoning segmentation dataset to fine-tune the model, enabling it to conduct professional dialogues and generate segmentation masks based on complex reasoning in the food domain. Our fine-tuned FoodLMM achieves state-of-the-art results across several food benchmarks. We will make our code, models and datasets publicly available.
\end{abstract}

\vspace{-0.8cm}
\section{Introduction}
\label{sec:intro}
\vspace{-0.2cm}
Benefiting from the remarkable language understanding capabilities of Large Language Models (LLMs)~\cite{ouyang2022training, brown2020language, wu2023brief, touvron2023llama, touvron2023llama2} and image features obtained from pretrained vision-language models~\cite{radford2021learning,li2022blip,li2023blip,fang2023eva}, Large Multi-modal Models (LMMs)~\cite{li2023blip, instructblip, liu2023visual, zhu2023minigpt, gong2023multimodal, wu2023next, chen2023minigpt} have exhibited outstanding performance in various vision-language tasks, such as image captioning, visual question answering and complex visual reasoning. Furthermore, LMMs are able to interact with humans with natural language,
paving the way to build versatile conversational assistants in different vertical domains~\cite{li2023llava}.
\par
Although previous LMMs perform well on general images and questions, due to a lack of domain-specific expertise, they often cannot provide reliable assistance in vertical domains and even produce incorrect responses or hallucinations~\cite{li2023llava}. This is particularly evident in the food realm.
For instance, when asked about nutrition in a food image, general LMMs typically can only answer what nutritional elements it contains but fail to provide specific quantities and precise nutritional content of the food. In food domain, the previous efforts have been devoted to several tasks, ranging from food and ingredient recognition~\cite{bossard2014food, chen2016deep, bolanos2017food, chen2020zero, zhang2022sequential, gao2022dynamic}, recipe generation~\cite{salvador2017learning, salvador2019inverse, h2020recipegpt, wang2020structure, wang2022learning}, nutritional estimation~\cite{thames2021nutrition5k} to food segmentation~\cite{freitas2020myfood, okamoto2021uec, wu2021large, zhu2023new, dong2021windows, honbu2022unseen, lan2023foodsam, sinha2023transferring}. 
Though the existing works achieve promising results in each individual task, they are incapable of dealing with other tasks using a single model.
\par

\begin{figure}[!t]
  \centering
  \includegraphics[width=1.0\textwidth]{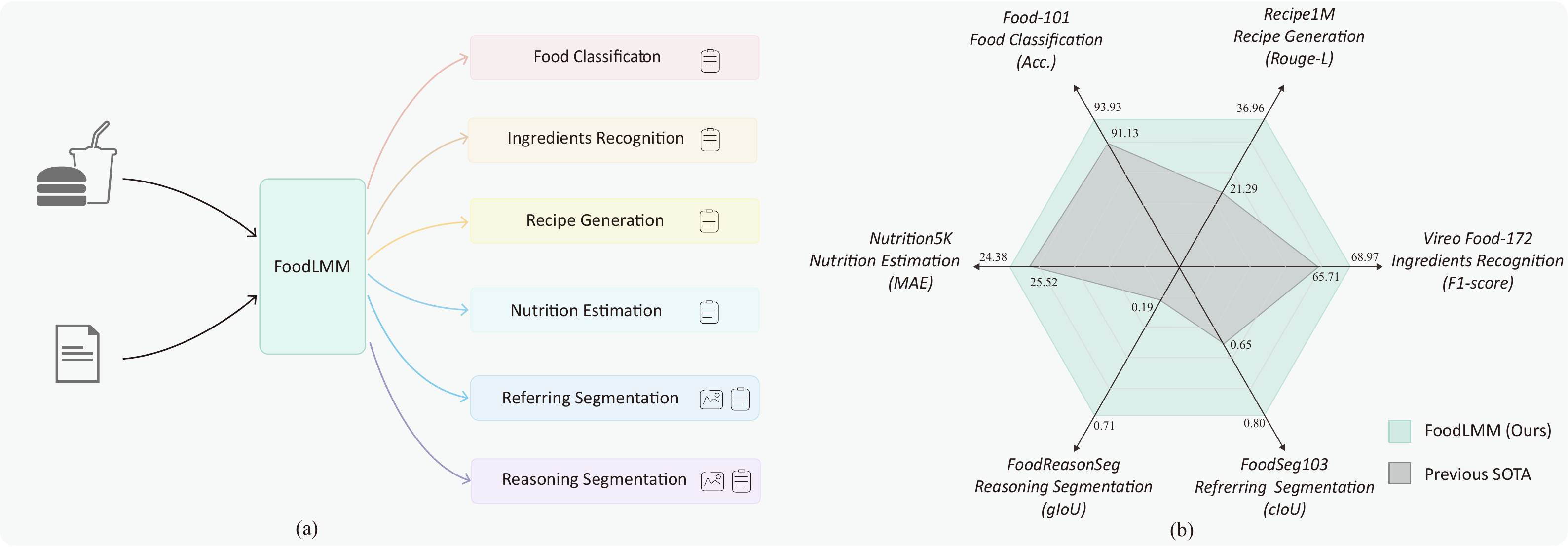}
  \caption{
  (a) FoodLMM is a versatile large multi-modal model for various food-related tasks.
  (b) Performance comparison between our FoodLMM and previous state-of-the-art (SOTA) methods across various tasks. 
  }
  \label{fig:intro}
\vspace{-0.6 cm}
\end{figure}

In this paper, we address the limitations by proposing FoodLMM, a versatile food assistant using a Large Multi-modal model tailored for various food-related tasks, As shown in Figure~\ref{fig:intro}. To the best of our knowledge, FoodLMM is the first unified and multitask LMM in the food domain.  
FoodLMM is capable of handling a variety of food-related tasks including: 
\textbf{Food Classification}, \textbf{Ingredient Recognition}, \textbf{Recipe Generation}, \textbf{Nutrition Estimation}, \textbf{Referring Segmentation}, and \textbf{Reasoning Segmentation},
achieving state-of-the-art (SOTA) performance in each task. 
The architecture of FoodLMM extends from LISA~\cite{lai2023lisa}, which consists of a base multi-modal large language model LLaVA~\cite{liu2023visual}, a segmentation model SAM~\cite{kirillov2023segment} and a series of novel task-specific head tokens.
Specifically, our FoodLMM produces text outputs for food classification, ingredient recognition and recipe generation tasks in an instruct-following fashion, similar to other LMMs~\cite{lai2023lisa, li2023llava}. Importantly,
we introduce multiple special segmentation tokens to the vocabulary and feed their hidden states to the segmentation decoder to generate one or multiple masks.
Notably, by employing this mask generation paradigm with our elaborate instructions, FoodLMM is empowered with the capability to solve the most challenging one-to-many and one-to-zero problems~\cite{hu2023beyond} in the \textbf{Referring Segmentation} task.
Additionally, a series of nutritional task-specific tokens are introduced for nutrition estimation, where their hidden states from the base LMM are processed through corresponding regression heads to yield precise values for various nutritional values. 

The training process of FoodLMM is divided into two stages. 
In the first stage, we utilize multiple public datasets~\cite{chen2016deep,salvador2017learning,thames2021nutrition5k,wu2021large,okamoto2021uec} to conduct multi-task learning, aiming at injecting substantial basic food domain knowledge into FoodLMM. Specifically, we design a rich set of instructions and answer query templates for each task,
enabling the model with the ability to handle different food tasks.
The second stage aims to endow the model with multi-round conversational interaction capabilities in free-form prompts. For this purpose, we create a food multi-round conversation dataset \textbf{FoodDialogues} and a reasoning segmentation dataset \textbf{FoodReasonSeg} with GPT-4~\cite{koubaa2023gpt}, for training. The FoodDialogues is constructed based on the Nutriton5k food nutrition dataset, where detailed nutritional information is fed to GPT-4, prompting it to generate multi-round dialogues on various food-related topics, such as calorie calculation, dietary planning, and metabolism. Similarly, FoodReasonSeg is derived from the FoodSeg103 food segmentation dataset, utilizing the GPT-4 to create multi-round complex reasoning dialogues, the masks of the mentioned ingredients as reasoning segmentation labels.
In the second training stage, these two datasets are added for fine-tuning. As a result, FoodLMM acquires the ability to interact in multi-round conversations with users and provide reasoning segmentation masks for complex queries.
Our main contributions can be summarized as follows:
\begin{itemize}
    \item We propose \textbf{FoodLMM}, a versatile large multi-modal model in the food domain. We design specific instructions for different tasks and unify a variety of food tasks using LMM, achieving SOTA performance across multiple benchmarks. To the best of our knowledge, our FoodLMM is the first unified approach for multiple food tasks. FoodLMM provides practical guidance for the construction of LMM in other vertical domains.
    \item FoodLMM generates segmentation masks and nutritional predictions through a series of novel task-specific tokens. This paradigm easily solves the most challenging one-to-many and one-to-zero problems in referring segmentation and achieves the ability to estimate the nutritional value of the entire dish or any ingredient in the food image.
    \item We construct diverse multi-round conversation food dataset  \textbf{FoodDialogues} and food
    reasoning segmentation dataset \textbf{FoodReasonSeg}.
    We will release both datasets, along with our codebase and model checkpoints to facilitate future research.
\end{itemize}

\vspace{-0.5cm}
\section{Related Work}
\label{sec:RW}
\vspace{-0.15cm}
\subsection{Large Multi-modal Model}
\par
The remarkable natural language capabilities of large language models (LLMs) lead researchers to explore their use in visual tasks. Recently, GPT-4~\cite{koubaa2023gpt} received significant recognition for introducing the ability to process multimodal inputs. Additionally, many multimodal LLMs that are available to the public have shown their effectiveness in general fields, such as Minigpt4~\cite{zhu2023minigpt}, Vicuna~\cite{chiang2023vicuna}, LLaVA~\cite{liu2023visual}, MultiModal-GPT~\cite{gong2023multimodal} and OpenFlamingo~\cite{awadalla2023openflamingo}.
There are successful cases of applying large language models in areas such as biomedicine~\cite{wu2023pmc, li2023chatdoctor, luo2023biomedgpt}, finance~\cite{wu2023bloomberggpt}, and law~\cite{cui2023chatlaw}, but these applications are confined to a single text modality.
LLaVA-Med~\cite{li2023llava} is a good specialization of LMM into the medical domain, which employs biomedical conversations generated by GPT-4 to instruct the model in learning biomedical knowledge.
However, there are few applications of LMMs in other domains, especially the food domain, which motivates us to propose FoodLMM.

\vspace{-0.2cm}
\subsection{Food Analysis}
With the release of food-related datasets, such as Food-101~\cite{bossard2014food}, VIREO Food-172 and 251~\cite{chen2016deep, chen2020study}, Recipe1M~\cite{salvador2017learning}, Nutrition5K~\cite{thames2021nutrition5k}, UEC Food256~\cite{kawano2015automatic}, FoodSeg103~\cite{wu2021large}, Food2K~\cite{min2023large}, the past efforts in food domain have been devoted to various tasks such as food classification~\cite{liu2016deepfood,ofli2017saki,kiourt2020deep,mezgec2017nutrinet,chen2017chinesefoodnet,martinel2018wide}, ingredient recognition~\cite{chen2016deep, bolanos2017food, chen2020study, chen2020zero, liu2020food}, cross-modal recipe retrieval~\cite{chen2017cross, zhu2019r2gan, chen2018deep,chen2017cross}, recipe generation~\cite{salvador2019inverse,h2020recipegpt,wang2020structure}, food segmentation~\cite{he2013food, ege2019new, wu2021large,hollywood2007using}, food recommendation~\cite{min2019food,elsweiler2017exploiting,freyne2010intelligent}, nutrition estimation~\cite{brown2001importance,thames2021nutrition5k,mirel2013national,chen2020national} and food logging~\cite{kitamura2008food,sahoo2019foodai,chen2019use}. 
These works focus on a specific food task. In contrast, this paper introduces a unified model FoodLMM to deal with various tasks.

\vspace{-0.2cm}
\subsection{Referring Segmentation}
Referring segmentation~\cite{hu2016segmentation} tasks aim to use instructions to guide the segmentation of specific objects mentioned in the query text. 
Transformer-based backbones are dominant in referring segmentation~\cite{ding2021vision, wang2022cris}.
SAM \cite{kirillov2023segment} attracts widespread attention in the community due to its precise segmentation and powerful zero-shot capabilities, but its performance on natural language prompts is suboptimal.
However, these referring segmentation studies assume a single object in the query text, not considering scenarios with no or multiple objects.
Considering this issue, ~\cite{hu2023beyond} proposes One-to-Many and One-to-Zero segmentation. 
LISA~\cite{lai2023lisa} introduces a novel reasoning segmentation task that demands complex query text. However, LISA rigidly provides a binary mask for each query.
In contrast, FoodLMM accurately conducts reasoning segmentation in the food domain, offering multiple masks or refraining from providing masks for non-existent objects.

\vspace{-0.2cm}
\section{Food Visual Instruction-Following Data}
\label{sec:dataset}
We follow LLaVA~\cite{liu2023visual} to adopt visual instruction tuning to train our FoodLMM.
Denote $\mathbf{X}_{\mathrm{i}}$ as a food image, $\mathbf{X}^m_{\mathrm{q}}$ and  $\mathbf{X}^m_{\mathrm{a}}$ as the textual query and answer of the $m$-th conversation turn respectively. The format of the instruction-following data is:
\begin{equation}
\label{eq:conversation_pattern}
\begin{aligned}
& \mathbf{X}_{\text{system-message}} \color{darkgreen}{\langle \text{STOP}\rangle}\color{black} \backslash \text{n} \\
&\text{USER: } \mathbf{X}_{\mathrm{i}}\mathbf{X}^1_{\mathrm{q}}\color{darkgreen}{\langle \text{STOP}\rangle}\color{black}   \backslash \text{n}
\text{ASSISTANT: } \color{darkgreen}{\mathbf{X}^1_{\mathrm{a}}\langle \text{STOP}\rangle}\color{black}  \backslash \text{n} \\
&\text{USER: } \mathbf{X}^2_{\mathrm{q}}\color{darkgreen}{\langle \text{STOP}\rangle}\color{black}   \backslash \text{n}
\text{ASSISTANT: } \color{darkgreen}{\mathbf{X}^2_{\mathrm{a}}\langle \text{STOP}\rangle}\color{black}  \backslash \text{n} \ldots
\end{aligned}
\nonumber
\end{equation}

Only \color{darkgreen}{green sequence/tokens} \color{black}are used to compute the auto-regressive loss.

FoodLMM's training process consists of two stages. 
Stage 1 aims to inject sufficient basic food knowledge into the LMM using various public food datasets. 
The conversational capability is empowered by our GPT-4 generated datasets in stage 2. 

\vspace{-0.2cm}
\subsection{Stage 1: Public Food Datasets}
\label{sec:data_stage1}
\vspace{-0.1cm}
We construct the instruction-following data based on instruction templates from $5$ most critical tasks in the food domain:  Food Classification, Ingredient Recognition, Recipe Generation, Nutrition Estimation and  Food Segmentation. The public datasets used for each task are listed in Table~\ref{tab:dataset_summary}, and we introduce them individually below.

\vspace{-0.5 cm}
\begin{table}[htbp]
\centering
\scalebox{1.0}{
\begin{tabular}{cccccc}
\toprule
Dataset    & Images     & Class & Ingredients & Recipes & Annotations   \\
\midrule
VIREO Food-172~\cite{chen2016deep} & 100k   & 172   & 353   & ——   & Category, ingredient \\
Recipe1M~\cite{salvador2017learning} & 626k & —— & 1,488 & 361k & Recipe \\
Nutrition5k~\cite{thames2021nutrition5k} & 125k  & —— & 555  & ——  & Ingredient, nutrition \\
FoodSeg103~\cite{wu2021large} & 7k & —— & 103  & ——  & Ingredient, mask  \\
UEC-FoodPIX~\cite{okamoto2021uec} & 10k & ——  & 102 & ——   & Ingredient, mask   \\
\bottomrule
\end{tabular}
}
\vspace{0.27cm}
\caption{Dataset Statistics of public food datasets to construct visual instruction-following data in stage 1.\label{tab:dataset_summary}}
\end{table}

\vspace{-1cm}

\noindent \textbf{Food VQA.}
We collectively refer to the three pure text output tasks: food classification, ingredient recognition and recipe generation as Food VQA and design instruction templates for each of them. We select two widely used benchmarks for Food VQA, VIREO Food-172~\cite{chen2016deep} and Recipe1M~\cite{salvador2017learning}. 
We only use recipes from the Recipe1M dataset that have corresponding images as cross-modal recipe retrieval~\cite{zhang2016multi,chen2016deep,chen2017cross}. 
We have elaborately designed 10, 8 and 11 instruction templates for food classification, ingredient recognition and recipe generation tasks respectively. Please refer to the supplementary materials for more details.

\noindent \textbf{Nutrition Estimation.}
Nutrition5k~\cite{thames2021nutrition5k} provides fine-grained nutritional element values, food calories and quality annotations, including RGB images, multi-angle videos and depth images.
We select RGB images and video frames with good viewing angles. Note that we do not use depth images, as they are difficult for users to acquire.
We design a rich collection of instruction templates for nutrient estimation tasks that contain 7 types of instructions, including queries about overall calories or nutritional values, inquiries referring to one or more ingredients, questions about the primary nutrient, etc. There are 64 query templates and 68 answer templates in total, listed in the supplementary material.

\noindent \textbf{Food Segmentation.}
The FoodSeg103~\cite{wu2021large} and UEC-FoodPIX Complete~\cite{okamoto2021uec} datasets are the two most commonly used benchmarks for food segmentation, containing masks of 103 and 102 ingredients respectively. FoodSeg103 annotates dishes at a more fine-grained level and proves to be a more challenging benchmark for food image segmentation~\cite{lan2023foodsam}. 
We design 10 query templates and 9 answer templates for food segmentation task.
All the instruction templates can be found in the supplementary material.

\begin{figure}[t!]
  \centering
    \includegraphics[width=1.\textwidth]{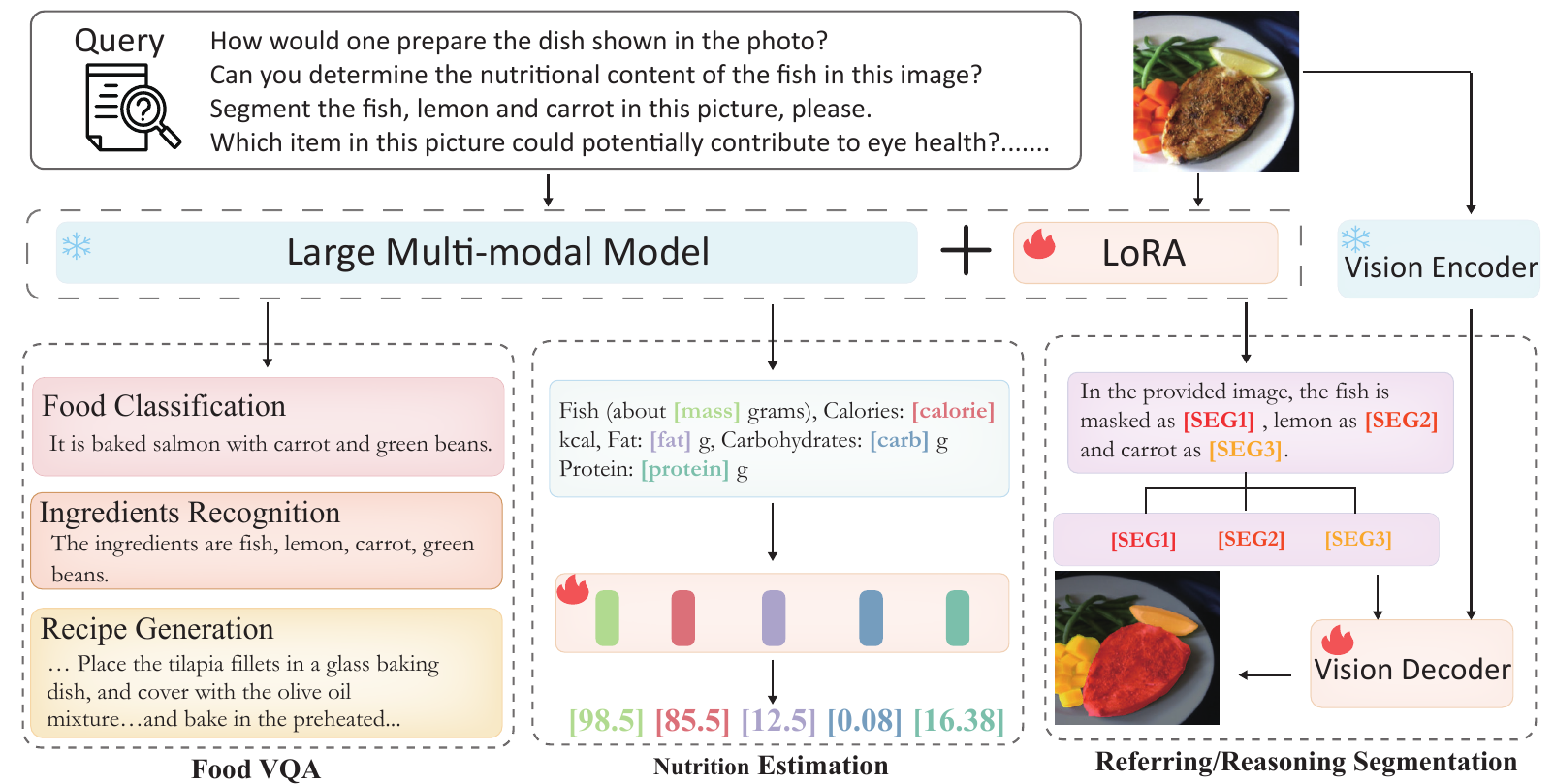} 
  \caption{Architecture Overview of the proposed FoodLMM.} 
  \label{fig:overview}
\vspace{-0.55 cm}
\end{figure}

\vspace{-0.2cm}
\subsection{Stage 2: GPT-4 Generated Conversation Datasets}
\vspace{-0.1cm}
Stage 2 aims to empower our FoodLMM with Multi-round conversational ability on various topics based on food images and provide segmentation masks for queries requiring complex reasoning. Nevertheless, the food domain lacks comprehensive instruction-following datasets and reasoning segmentation datasets suitable for LMMs.
To bridge this gap, as shown in Table~\ref{tab:FoodDialogues}, we construct the first food Multi-round conversation dataset \textbf{FoodDialogues} and food reasoning segmentation dataset \textbf{FoodReasonSeg}.

\textbf{FoodDialogues} is built from the Nutrition5k dataset, which contains ingredient labels and precise nutrition information, making it unique and suitable for various conversational topics. 
Specifically, we follow the training and testing splits of the original data set and selected an overhead RGB image and a well-angled video frame for each sample. Send the sample's ingredient list and detailed nutritional information to GPT-4 in the form of plain text, and request GPT-4 to generate multiple rounds of conversations on different topics, ranging from nutrition, calorie calculation, health and diseases, metabolism, dietary planning, allergies, food pairing to substitution. 
The prompt for GPT-4 is carefully designed based on expert opinions where answers are required to be professional and explanatory.

\textbf{FoodReasonSeg} is constructed based on the food segmentation dataset FoodSeg103. Similarly, we send the ingredients list to GPT-4 and ask it to generate multi-round conversations where the questions should require complex reasoning. The corresponding masks provided in the original dataset for the ingredients mentioned in answers are used as segmentation labels. 

Moreover, we conduct a human evaluation to measure the quality of the generated \textbf{FoodDialogues} and \textbf{FoodReasonSeg} datasets. Specifically, users are asked to rate the quality of GPT-4 generated content on a five-point scale (5 being the best).
The average score is 4.2 for FoodDialogues and 4.4 for FoodReasonSeg, with more than half of the participants choosing 5, which showcases the remarkable reliability and high quality of the datasets.
More details are provided in the supplementary material.

\vspace{-0.5 cm}
\begin{table}[htbp]
\centering
\begin{tabular}{ccccc}
\toprule
Dataset & Split & Images & Dialogues & QA Pairs \\
\midrule
\multirow{2}{*}{FoodDialogues}& Train & 8,094    & 6,428   & 25,821        \\
& Test &  1,418   & 1,094         & 4,325       \\
\midrule
\multirow{2}{*}{FoodReasonSeg}& Train & 3,907    & 3,997        & 13,817                     \\
& Test &  1,620   & 1,703         & 5,831                     \\
\bottomrule
\end{tabular}
\vspace{0.27cm}
\caption{Dataset Statistics of FoodDialogues and FoodReasonSeg.\label{tab:FoodDialogues}}
\vspace{-0.3 cm}
\end{table}
\vspace{-1 cm}

\vspace{-0.2cm}
\section{Method}
\label{sec:method}
The network architecture of FoodLMM is illustrated in Figure~\ref{fig:overview}. 
Our FoodLMM takes image and text prompts as input, and is capable of handling various food tasks in a united model. The architecture of FoodLMM is built upon Large Multi-modal Model (LMM) LISA~\cite{lai2023lisa}, 
which is LMM with powerful language generation and image segmentation ability. 
As we not only focus on language generation tasks, such as ingredient recognition and recipe generation but also food segmentation, we naturally choose LISA instead of other LMMs, such as LLaVA~\cite{li2023llava}, to build our FoodLMM. 

We adopted a two-stage training strategy to specialize the general LMM in the food domain.
The first training stage aims to inject sufficient food-related basic knowledge into the LMM by using instruction-following data constructed in~\ref{sec:data_stage1} based on five tasks in the food domain. As the output of the three tasks of food classification, ingredient recognition and recipe generation are pure text, these tasks could be learned through pure language auto-regressive similar to LLaVa. 
The output of food segmentation is multi-modal, with both masks and text, and nutrition estimation aim to produce nutrition values e.g., mass and calorie. We propose task-specific tokens and heads for food segmentation and nutrition estimation to address this issue.
The segmentation masks are generated through multiple additional segmentation tokens and the segmentation encoder-decoder.
Different types of nutritional values are predicted by a series of nutrition tokens and regression heads. 
The LMM is fine-tuned using LoRA to learn the generation of task-specific tokens and the answers for various tasks.
The second training stage focuses on endowing FoodLMM with the capability to conduct multiple rounds of highly professional conversations and provide detailed explanations with segmentation masks for queries requiring complex reasoning. 

\vspace{-0.2cm}
\subsection{Stage 1: Multi-task learning}
\label{sec41}
We adopt a multi-task learning approach in this stage, in order to enable LMM to handle different basic food tasks. 
The tasks include \textbf{Food VQA} (food classification,  ingredient recognition, recipe generation),  \textbf{Nutrition Estimation} and \textbf{Referring Segmentation}.

\noindent \textbf{Food VQA.} We design different question-and-answer templates for food classification, ingredient recognition and recipe generation tasks. Similar to LLaVA, FoodLMM is trained by language autoregression. The loss can be formalized as follows:
\begin{equation}
\label{eq:generation_loss}
\mathcal{L}_{txt}=\mathbf{CE}(\hat{\boldsymbol{y}}_{txt}, \boldsymbol{y}_{txt}),
\end{equation}
where $\hat{\boldsymbol{y}}_{txt}$ stands for the generated sentences (tokens), $\boldsymbol{y}_{txt}$ represents for the ground-truth. 

\noindent \textbf{Nutrition Estimation.}
We introduce nutritional tokens in FoodLMM for nutrition estimation.
Specifically, a total of ten nutritional tokens are added into the vocabulary: $\langle mass \rangle$, $\langle cal \rangle$, $\langle carb \rangle$, $\langle fat \rangle$, $\langle pro \rangle$ are used for regressing the mass, calories, carbohydrates, fats, and proteins of any ingredient respectively,
and $\langle total\_mass \rangle$, $\langle total\_cal \rangle$, $\langle total\_carb \rangle$, $\langle total\_fat \rangle$, $\langle total\_pro \rangle$ are used for predicting the overall nutritional element values of the input food picture.

The hidden states of these nutritional tokens are processed into embeddings through a Multilayer Perceptron (MLP). These embeddings are fed into the corresponding regression heads to get the predicted values. Finally, the predicted values replace the task-specific tokens in the original text output to obtain the answer with nutritional value.

We employ the Mean Absolute Error (MAE) and the Mean Squared Error Loss (MSE) as follows:
\begin{equation}
\label{eq:nutrition_loss}
\mathcal{L}_{\text {nutrition }}=\lambda_{\text {MAE}}(\frac{1}{n} \sum_{i=1}^{n}\left|y_{i}-\hat{y}_{i}\right|)+\lambda_{\text {MSE}}(\frac{1}{n} \sum_{i=1}^{n}\left(y_{i}-\hat{y}_{i}\right)^{2}) ,
\end{equation}
where $\hat{y}_{i}$ stands for the predictions, $y_{i}$ for the ground-truth values, $i$ for different regression heads,
$\lambda_{\text {MAE}}$ and $\lambda_{\text {MSE}}$ are the corresponding weights of the MAE and MSE losses. 

\noindent \textbf{Referring Segmentation.} To embed ingredient segmentation into our FoodLMM, we introduce segmentation tokens, denoted as $\langle seg_i\rangle$, with $i$ representing the ingredient's index, into the primary vocabulary. When the model receives a query and its associated image, it generates a text response containing tokens for the ingredients to be segmented. The hidden states of these tokens are converted through an MLP into embeddings that capture the relevant ingredient information. These embeddings, combined with the visual representation from SAM's encoder, are processed by the SAM decoder to create the ingredient masks. 

The segmentation loss is denoted as $\mathcal{L}_{mask}$, integrating the per-pixel binary cross-entropy (BCE) loss alongside the Dice coefficient loss (DICE) ~\cite{milletari2016v} as follows:
\begin{equation}
\label{eq:mask_loss}
\small{
\mathcal{L}_{mask} = \lambda_{bce}\frac{1}{n} \sum_{i=1}^{n} \operatorname{BCE}(\hat{M_i}, M_i) + \lambda_{dice} \frac{1}{n} \sum_{i=1}^{n}\operatorname{DICE}(\hat{M_i}, M_i)}
\end{equation}
where $\hat{M_i}$ and $M_i$ stand for the predicted mask and the ground-truth of token $\langle seg_i\rangle$ respectively.  
Every projected embedding of $\langle seg_i\rangle$ token is fed into the decoder successively to generate the corresponding $\hat{M_i}$.

Almost all the referring segmentation models are limited to producing a single binary mask for different queries~\cite{hu2016segmentation,li2018referring,margffoy2018dynamic,yang2022lavt}.
To address this issue, we aim to enable FoodLMM to go \textit{Beyond One-to-One}~\cite{hu2023beyond}, i.e., to produce different numbers of masks based on the queries.
To achieve this, we design three different types of referring instructions: (1) to segment certain specified ingredients (one-to-one/one-to-many), e.g., Query: Segment the fish and lemon in this picture. Answer: The fish is masked as $\langle seg_1\rangle$ and the lemon as $\langle seg_2\rangle$. (2) to segment all visible ingredients (one-to-many), e.g., Query: Segment all the ingredients in this photo. Answer: The fish is masked as $\langle seg_1\rangle$, lemon as $\langle seg_2\rangle$, carrot as $\langle seg_3\rangle$ and green beans as $\langle seg_4\rangle$. (3) to segment objects that are not present in the image (one-to-zero), e.g., Query: Segment the watermelon in this picture. Answer: The watermelon is not found in this picture. Through these three instructions, the challenge of referring segmentation is dexterously addressed, realizing \textit{One-to-Any} segmentation. 

Our FoodLMM is trained by combining all the tasks and data with multitask learning. Thus, the overall loss $\mathcal{L}$ consists of $\mathcal{L}_{txt}$, $\mathcal{L}_{mask}$ and $\mathcal{L}_{nutrition}$, weighted by $\lambda_{txt}$, $\lambda_{nutrition}$ and $\lambda_{mask}$, as formulated below:
\begin{equation}
\label{eq:total_loss}
\mathcal{L}=\lambda_{txt}\mathcal{L}_{txt}+\lambda_{nutrition}\mathcal{L}_{nutrition} +\lambda_{mask}\mathcal{L}_{mask}.
\end{equation}

\vspace{-0.5cm}
\subsection{Stage 2:  Fine-tuning for a Versatile Conversational Food Assistant}
\label{sec:stage2}
The second stage focuses on training FoodLMM's \textbf{Multi-round Conversation} and \textbf{Reasoning Segmentation} capabilities. Relying on the generated datasets \textbf{FoodDialogues} and \textbf{FoodReasonSeg}, a versatile food conversational assistant can be achieved. 

\noindent \textbf{Multi-round Conversation.}
We adjust the generated dialogues to the format of instruction-following data (defined in Section~\ref{sec:dataset}) to train FoodLMM's multi-round conversation capability. 
In one conversation, the image is input only before the first round of questioning. 
In subsequent rounds, each new question is appended to the existing historical context, guiding the LMM to produce a subsequent response. The associated loss function is the autoregression loss $\mathcal{L}_{txt}$ (Eq~\ref{eq:generation_loss}).

\noindent \textbf{Reasoning Segmentation.}
The data format is consistent with multi-round conversation. Every answer contains segmentation tokens $\langle seg_i\rangle$, and segmentation loss $\mathcal{L}_{mask}$ (Eq~\ref{eq:mask_loss}) is used to enhance the segmentation capability.

\vspace{-0.2cm}
\section{Experiment}
\label{sec:experiment}
\vspace{-0.15cm}
\begin{figure*}
  \centering
    \includegraphics[width=1.0\textwidth]{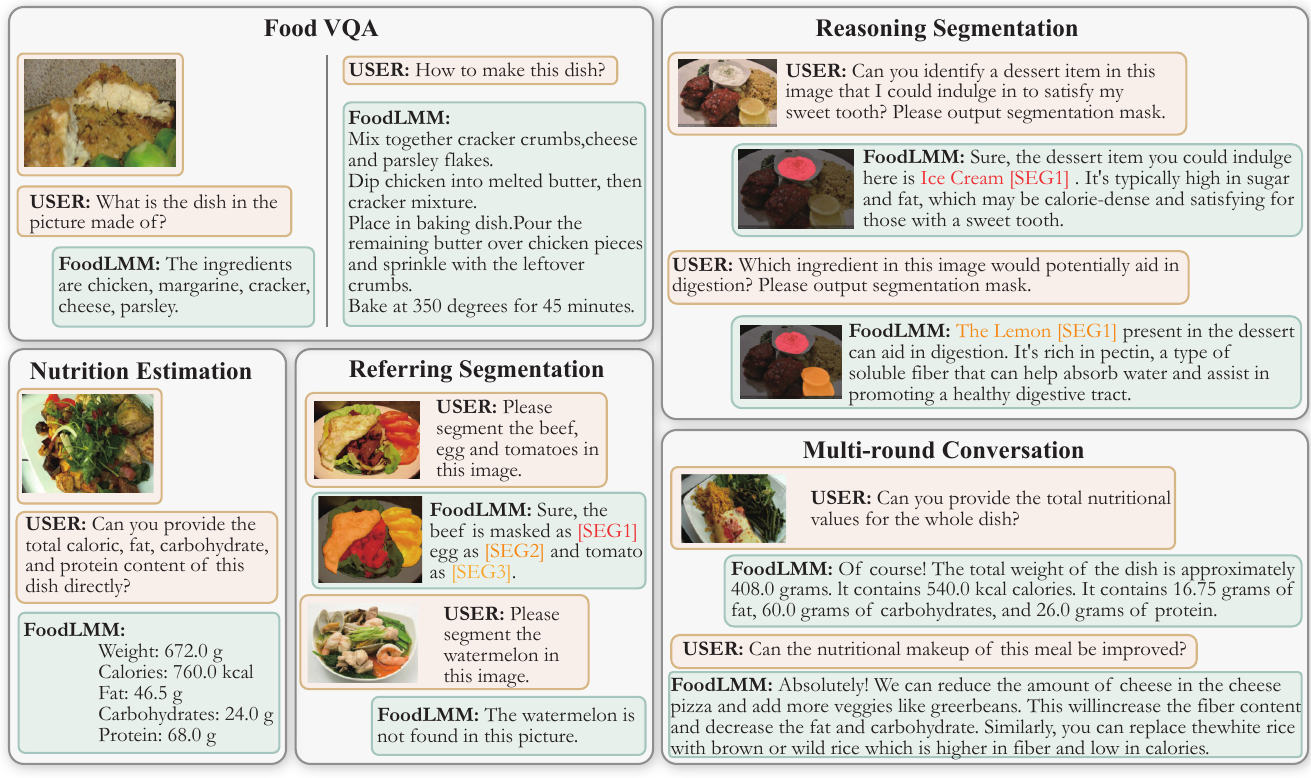} 
  \caption{Qualitative examples of various capabilities of FoodLMM.}
  \label{fig:tasks}
\vspace{-0.5 cm}
\end{figure*}

\subsection{Experimental Setup}
\textbf{Training Details.} We train our FoodLMM from LISA-7B-v1-explanatory model~\cite{lai2023lisa}, using four NVIDIA 40G A100 GPUs. AdamW optimizer combined with the WarmupDecayLR is used as a learning rate scheduler. We set the initial learning rate to 0.0003, the weight decay to 0, and used 100 warm-up iterations. 
Following LISA, We set the weights $\lambda_{BCE}$ and $\lambda_{Dice}$ to 2.0 and 0.5 respectively.
For the nutrition estimation task, we empirically adjust the weights $\lambda_{MAE}$ and $\lambda_{MSE}$ to 0.1 and 0.0001 respectively to balance their magnitudes with other losses.
Throughout the training, we employed LoRA to fine-tune LLaVa. In addition, the parameters for SAM's decoder, the MLP, and the heads dedicated to nutritional value estimation are all trainable. 

In stage 1, the model is trained with a batch size of 4 and the sampling rate of three tasks, i.e., Food VQA, Nutrition Estimation and Food Segmentation is set to $2:1:1$. We denote the model trained from Stage 1 as \textbf{FoodLMM S1}. In Stage 2, the batch size is reduced to $2$, and the sampling ratio of Multi-turn Conversation and Reasoning Segmentation is set to $3:2$.
To retain the capabilities of the tasks in the first stage, we replayed the training data of the first stage, and the total sampling ratio of the generated data to the public data is set to $15:7$.
The model trained from Stage 2 is denoted as \textbf{FoodLMM Chat}.
To better evaluate the ability of FoodLMM on specific tasks, we fine-tuned FoodLMM on each task, recorded as \textbf{FoodLMM FT}, and the batch size during fine-tuning is set to 4.

\vspace{-0.2cm}
\subsection{Evaluation Metrics.}
For \textbf{food classification}, top-1 accuracy is adopted as the evaluation metric. While for \textbf{ingredient recognition}, the Intersection over Union (IoU) and F1-score are used to evaluate the performances. For \textbf{recipe generation}, we follow  \cite{salvador2019inverse, chhikara2023fire} and apply SacreBLEU, Rouge-L metrics to quantify the quality of the generated recipes. For 
\textbf{nutrition estimation}, following~\cite{thames2021nutrition5k}, we use mean absolute error (MAE) and the percent of MAE to the respective mean for that field to measure regression accuracy for calories, mass, and individual macronutrient mass. Caloric MAE is measured in kilocalories, all others are measured in grams.
For \textbf{referring segmentation}, in line with LISA~\cite{lai2023lisa}, we employ the Complete Intersection over Union (cIoU) metric to evaluate the performance. Meanwhile, to evaluate the risk of our model incorrectly rejecting segmenting existing ingredients, we also report the probability of accurate response. For one-to-zero referring segmentation, we report the accuracy of our model successfully rejecting to segment non-existent items.
For \textbf{reasoning gegmentation}, following LISA~\cite{lai2023lisa}, Generalized Intersection over Union (gIoU) along with cIoU are adopted for the evaluation of Reasoning Segmentation. 

\begin{table}[!t]
\centering
\scalebox{0.54}{
\begin{tabular}{cccccccc}
\toprule
\multicolumn{2}{c}{Food Classification in Food-101~\cite{bossard2014food}} & \multicolumn{3}{c}{Ingredient Recognition in VIREO Food-172~\cite{chen2016deep}}  & \multicolumn{3}{c}{Recipe Generation in Recipe1M~\cite{salvador2017learning}} \\
Method & Acc. & Method & IoU & F1 & Method & SacreBLEU & Rouge-L \\
PRENet~\cite{min2023large} & 91.13 & CACLNet~\cite{luo2023ingredient} & —— & 65.71 & FIRE~\cite{chhikara2023fire} & 6.02 & 21.29 \\
FoodLMM FT & \textbf{93.93} & FoodLMM FT & \textbf{56.94} & \textbf{68.97} & FoodLMM FT* & \textbf{6.24} & \textbf{36.96}\\
\midrule
\multicolumn{8}{c}{Nutrition Estimation}  \\
Dataset & Method  & Total Caloric MAE & Total Mass MAE & Total Fat MAE & Total Carb MAE & Total Protein MAE & Average \\
& 2D Direct Prediction~\cite{thames2021nutrition5k} & 70.6 / 26.1\% & 40.4 / 18.8\% & 5.0 / 34.2\% & 6.1 / 31.9\%  & 5.5 / 29.5\% & 25.52 / \textbf{28.1\%}  \\
& FoodLMM S1 & 67.3 / 26.6 \%  & 38.8 / 20.2 \%  & 5.5 / 40.4 \%  & 6.1 / 31.9 \%  & 4.2 / 26.2 \%  & \textbf{24.38} / 29.1\%  \\
\multirow{-3}{*}{Nutrition5k~\cite{thames2021nutrition5k}} & FoodLMM FT & 67.3 / 26.6 \% & 39.7 / 20.7 \%  & 5.4 / 39.7 \% & 5.9 / 31.1 \%  & 4.1 / 25.8 \%  & 24.48  / 28.8\% \\
\midrule
\multicolumn{8}{c}{Referring Nutrition Estimation} \\
Dataset & Refer ingredient & Model & Caloric MAE & Mass MAE & Fat MAE & Carb MAE & Protein MAE \\
&   & FoodLMM S1 & 69.3 / 45.8 \% & 56.1 / 43.9 \%  & 3.8 / 47.6 \%  & \textbf{3.2 / 30.3 \%} & 3.1 / 29.4 \% \\
& \multirow{-2}{*}{refer@1st} & FoodLMM FT & \textbf{25.2 / 34.7 \%}  & \textbf{47.3 / 37.1 \%}  & \textbf{3.7 / 46.1} \%  & 3.7 / 34.1 \%  & \textbf{2.5 / 22.4 \%} \\
&  & FoodLMM S1  & 34.8 / 45.5 \% & 21.2 / 34.1 \%  & \textbf{1.1} / 29.4 \% & \textbf{2.0 / 29.8 \%}  & 1.7 / 33.4 \% \\
& \multirow{-2}{*}{refer@2nd} & FoodLMM FT & \textbf{21.0 / 27.4 \%} & \textbf{16.9 / 27.1 \%}  & \textbf{1.1 / 29.0 \%}  & 2.1 / 30.5 \% & \textbf{1.4 / 27.4 \%}  \\
&  & FoodLMM S1 & 30.5 / 71.2 \%  & 21.0 / 55.4 \% & 1.1 / 50.5 \%  & 1.7 / 45.2 \%  & 1.6 / 66.7 \%  \\
\multirow{-6}{*}{Nutrition5k~\cite{thames2021nutrition5k}} & \multirow{-2}{*}{refer@3rd} & FoodLMM FT  & \textbf{16.9 / 39.5 \%}  & \textbf{14.2 / 37.5 \%} & \textbf{0.9 / 39.9 \%} & \textbf{1.5 / 38.1 \%} & \textbf{1.2 / 48.2 \%}  \\
\midrule
\multicolumn{8}{c}{Referring Segmentation (one-to-one/one-to-many)} \\
& & \multicolumn{2}{c}{refer@1} & \multicolumn{2}{c}{refer@3} & \multicolumn{2}{c}{refer@5}\\
\multirow{-2}{*}{Dataset} & \multirow{-2}{*}{Method}  & cIoU  & Acc. & cIoU  & Acc. & cIoU & Acc. \\
& LISA-7B~\cite{lai2023lisa} & 0.65 & 100.00\% & —— & —— & —— & —— \\
& FoodLMM S1 & 0.78 & 99.16\% & 0.73 & 98.99\%  & 0.58 & 98.34\%  \\
\multirow{-3}{*}{FoodSeg103~\cite{wu2021large}} & FoodLMM FT & \textbf{0.80} & 98.69\%  & \textbf{0.74}  & 99.51\% & \textbf{0.62} & 99.29\% \\
& LISA-7B~\cite{lai2023lisa} & 0.72 & 100.00\%  & —— & —— & —— & —— \\
& FoodLMM S1 & 0.90 & 99.70\%  & 0.72  & 97.51\% & 0.64 & 98.10\% \\
\multirow{-3}{*}{UEC-FoodPIX~\cite{okamoto2021uec}} & FoodLMM FT & \textbf{0.90} & 99.60\% & \textbf{0.76} & 99.38\% & \textbf{0.68} & 99.05\% \\
\midrule
\multicolumn{8}{c}{Referring Segmentation (one-to-zero)} \\
& & \multicolumn{2}{c}{Acc.} & & & \multicolumn{2}{c}{Acc.} \\
\multirow{-2}{*}{Dataset} & \multirow{-2}{*}{Method} & refer@1 & refer@3 & \multirow{-2}{*}{Dataset} & \multirow{-2}{*}{Method} & refer@1 & refer@3 \\
& LISA-7B & 0.00\% & —— & & \multicolumn{1}{l}{LISA-7B} & 0.00\% & —— \\
& FoodLMM S1 & 73.07\%  & 57.80\% & & \multicolumn{1}{l}{FoodLMM S1} & 77.50\% & 52.30\% \\
\multirow{-3}{*}{FoodSeg103~\cite{wu2021large}} & FoodLMM FT & \textbf{87.87\%}  & \textbf{88.99\%}  & \multirow{-3}{*}{UEC-FoodPIX~\cite{okamoto2021uec}} & \multicolumn{1}{l}{FoodLMM FT} & \textbf{83.70\%} & \textbf{83.50\%} \\
\bottomrule
\end{tabular}
}
\vspace{0.27cm}
\caption{Performance comparison on different food benchmarks. ``FoodLMM S1" denotes the model trained from the first stage, and ``FoodLMM FT" represents the model finetuned on specific tasks. 
} 
\label{tab:multi_task_results}
\vspace{-0.8 cm}
\end{table}

\vspace{-0.2cm}
\subsection{FoodVQA Results}
Our FoodVQA consists of three tasks, i.e., 
food classification, ingredient recognition, and recipe generation. For these tasks, we compare the performance of our model with baseline models that achieve state-of-the-art performances.
Note that other LMMs (e.g., LISA~\cite{lai2023lisa}) are general-purpose with limited knowledge in the food domain, their outputs of the FoodVQA tasks are uncontrollable and difficult to be quantitatively evaluated. Empirically, the results are also not promising. Consequently, we do not include the results of other LMMs as baselines for the FoodVQA tasks.
Table~\ref{tab:multi_task_results} summarizes the results. As shown in the table, for all these three tasks, our FoodLMM outperforms the SOTA methods and achieves the best results. Specifically, for food classification, by fine-tuning on Food-101 dataset, our model attains an accuracy of 93.93\%, which is 3.07\% higher than the SOTA method~\cite{min2023large}. For ingredient recognition, our model outperforms CACLNet~\cite{luo2023ingredient}, a SOTA method on VIREO Food-172 dataset, for 3.2\% in terms of F1. For recipe generation, without any extra information, our FoodLMM significantly outperforms FIRE \cite{chhikara2023fire}, the SOTA method on Recipe1M dataset that leverages ground-truth ingredient labels for recipe generation. Figure~\ref{fig:tasks}) further shows qualitative examples for foodVQA. As can be seen, our FoodLMM can generate accurate answers for food classification, ingredient recognition as well as recipe generation, showing the effectiveness of our model. 

\vspace{-0.2cm}
\subsection{Nutrition Estimation Results}
We further evaluate the performance of our model in nutrition estimation, the results are shown in Table~\ref{tab:multi_task_results}. 
After training Stage 1,  FoodLMM has been able to accurately estimate the overall nutritional value based on food images, decreasing the prediction error by $4.5\%$ on average compared to the previous SOTA~\cite{thames2021nutrition5k}.
Examples of using FoodLMM to estimate the total nutritional value of food are shown in Figure~\ref{fig:tasks}.
Moreover, FoodLMM is also able to predict the nutritional value of specific ingredients in dishes, which is named Referring Nutrition Estimation. It is worthwhile mentioning that previous methods lack the ability of Referring Nutrition Estimation. For evaluating the performance of Referring Nutrition Estimation, we select the top three ingredients with the largest mass as referred ingredients. Table~\ref{tab:multi_task_results} summarizes the results. In the table, refer@$k$st denotes referring the ingredient with the $k$th highest mass. The results show that although further fine-tuning does not enhance the performance of overall nutrition estimation, it significantly improves the accuracy of nutrition estimation for specific ingredients.

\vspace{-0.2cm}
\subsection{Referring Segmentation Results}
We compare our FoodLMM with LISA~\cite{lai2023lisa} on FoodSeg103 and UECFoodPIX datasets, the results are shown in Table~\ref{tab:multi_task_results}. For standard referring segmentation (one-to-one), FoodLMM's cIoU scores significantly exceed those of LISA, achieving 20\% improvement (0.65 to 0.78) on FoodSeg103 and 25\% (0.72 to 0.90) on UECFoodPIX datasets. 
Since we deliberately train FoodLMM's one-to-zero ability, the model may inevitably give a zero mask to the correct referring, i.e., mistakenly identifying the queried ingredient as non-existent. We examine the probability of accurate response (ACC) to measure this impact.
Notably, the decrease in FoodLMM's accuracy is minimal (99.16\% on FoodSEG103 and 98.69\% on UECFoodPIX) despite significant improvements in cIoU. 

\noindent\textbf{One-to-many.} We then evaluate the performance of segmentation involving multiple ingredients referred in a query. Table ~\ref{tab:multi_task_results} summarizes the segmentation results of three ($refer@3$) and five ($refer@5$) referred ingredients. 
The results show that FoodLMM, after the first training stage (FoodLMM S1), achieves high cIoU scores (around 0.7) on the $refer@3$ task on two datasets, while also maintaining high accuracy. However, as the number of referred ingredients increased
the segmentation become more challenging. Nevertheless, the fine-tuned model (FoodLMM FT) attains higher cIoU scores across all scenarios and maintains a high probability of accurate responses (over 99\%) even in complex segmentation tasks.
Figure~\ref{fig:tasks} shows the examples of referred ingredient segmentation. 

\noindent\textbf{One-to-zero}
To further examine the reasoning performance of our FoodLMM, we evaluate its performance under a challenging scenario, where the model is required to segment the ingredients that are not present in the image. For such a query, the model should reject to return the segmentation mask. The performances of one-to-zero segmentation are shown in Table~\ref{tab:multi_task_results}. As LISA is not trained with one-to-zero samples, the accuracy in identifying absent ingredients on both FoodSeg103 and UECFoodPIX datasets is 0.0\%. In contrast, our FoodLMM, attain an accuracy of 73.07\% on FoodSeg103 and 77.50\% on UECFoodPIX, when handling one single nonexistent referring ($refer@1$), after the first train stage. The task becomes more challenging when there are three absent referring (refer@3) in the query. Nevertheless, FoodLMM still maintains a high accuracy, surpassing 80\% after fine-tuning. Figure~\ref{fig:tasks} presents an example where the user requests FoodLMM to segment watermelon from the seafood vegetable soup, but the model declines.

\vspace{-0.2cm}
\subsection{Performance on the proposed Benchmarks}
\noindent \textbf{Multi-turn Conversation.}
We further evaluate the performance of our model on the Multi-turn Conversation dataset. The evaluation is performed by user study. Specifically, we invited 10 participants to manually evaluate the satisfaction score (1 to 5) of 20 sets of multi-turn dialogues (totalling 200 dialogue sets, all of the questions are randomly selected from the test set of our FoodDialogues). The results are summarized in Table~\ref{tab:S2_conversation}. The majority of the dialogues (173 out of 200) receive high scores ($>$=3), and 4 is the most frequent score. It indicates that our FoodLMM is capable of closely aligning with user preferences and generating high-quality answers that are both accurate and explanatory.

\vspace{-0.3 cm}
\begin{table}[htbp]
\centering
\begin{tabular}{cccccc}
\toprule
Score        & 1 & 2 & 3 & 4 & 5 \\
\midrule
Number       & 5 & 22 & 50 & 73 & 50  \\
\bottomrule
\end{tabular}
\vspace{0.27cm}
\caption{The human evaluation results of FoodLMM Chat.}
\vspace{-0.15cm}
\label{tab:S2_conversation}
\end{table}

\vspace{-1.3 cm}
\begin{table}[htbp]
\centering
\begin{tabular}{ccc}
\toprule
Method        & gIoU & cIoU \\
\midrule
LISA-7B       & 0.19 & 0.17 \\
FoodLMM S1    & 0.35 & 0.34 \\
FoodLMM Chat & 0.71 & 0.72 \\
\bottomrule
\end{tabular}
\vspace{0.27cm}
\caption{Performance comparison in reasoning segmentation.}

\label{tab:S2_seg}
\vspace{-0.7cm}
\end{table}

\vspace{-0.2cm}
\noindent \textbf{Reasoning Segmentation.}
We compare the performance of our FoodLMM with LISA in reasoning segmentation on our FoodReasonSeg benchmark. The results are shown in Table~\ref{tab:S2_seg}. From the results, LISA achieves a quite low gIoU and cIoU, which are $0.19$ and $0.17$, respectively. It indicates that although LISA is excellent in the general domain's reasoning segmentation, due to lacking specialized knowledge in the food domain, its performance in food reasoning segmentation is less impressive.
In contrast, our FoodLMM S1 achieves much better performance by learning food knowledge from public datasets. 
After the Stage 2 fine-tuning, the performance of FoodLMM is significantly improved, with the gIoU score increasing from $0.35$ to $0.71$ and the cIoU score from $0.34$ to $0.72$ compared to FoodLMM S1, indicates that FoodLMM acquired more expertise and stronger reasoning ability from the constructed datasets.

\subsection{Discussion about Multi-task and Single-task learning}
\vspace{-0.3cm}
\begin{table}[h]
\vspace{-0.2cm}
\centering
\scalebox{0.8}{
\begin{tabular}{cc|cc}
\toprule
\multicolumn{2}{c|}{Referring Segmentation One-to-One/Many (cIoU/Acc)} & \multicolumn{2}{c}{Referring Segmentation One-to-Zero (Acc)} \\
Multitask & 0.75 / \textbf{99.25\%} & Multitask & 86.02\%\\
Single-task & \textbf{0.76} / 98.20\% & Single-task & \textbf{89.07\%} \\
\midrule
\multicolumn{2}{c|}{Nutrition Estimation (MAE)} & \multicolumn{2}{c}{Referring Nutrition Estimation (MAE)} \\
Multitask & \textbf{24.48} / \textbf{28.8\%} & Multitask & \textbf{10.64} / \textbf{34.60\%} \\
Single-task & 25.6 / 30.5\% & Single-task & 16.40 / 43.62\% \\
\midrule
\multicolumn{2}{c|}{Ingredient Recognition (IoU/F1)} & \multicolumn{2}{c}{Recipe Generation (BLEU/Rouge-L)} \\
Multitask & \textbf{56.94} / \textbf{68.97} & Multitask & 5.89 / 35.12 \\
Single-task & 56.49	/ 68.72 & Single-task & \textbf{6.24} / \textbf{36.65} \\
\bottomrule
\end{tabular}
}
\vspace{0.27cm}
\caption{Performance comparison between multi-task learning and single-task learning. The average results are reported.}
\label{tab:multi-task}
\vspace{-0.8cm}
\end{table}

We compare the performance between single-task and multi-task learning of our FoodLMM in this section. Multi-task model is capable of dealing with various food tasks, while single-task model focuses on a specific task. Table~\ref{tab:multi-task} lists the performance differences between multi-task and single-task models. For the Referring Segmentation task, we present average segmentation results for variable number of ingredients across FoodSeg103~\cite{wu2021large} and UEC-FoodPIX Complete~\cite{okamoto2021uec}. For Nutrition Estimation task, we report the average prediction results for five essential nutrients. We observe that the advantage of multi-task is the ability to leverage knowledge across different tasks, which benefits some basic recognition tasks such as ingredient recognition and nutrition estimation. For more complex tasks, multi-task is only able to boost the performance of referring nutrition estimation. In general, single-task learning may benefit complex tasks such as recipe generation as model training can be more easily optimized. Nevertheless, multi-task learning still suffers from the problem of gradient conflicts between tasks~\cite{liu2019loss, yu2020gradient, crawshaw2020multi}, which remains an open issue and is worth further research.

\section{Conclusion}
We have presented FoodLMM, a versatile large multi-modal model for the food domain. This model is proficient in understanding and responding to various food-related questions, including Food Classification, Ingredient Recognition, Recipe Generation, Nutrition Estimation, Referring Segmentation, and Reasoning Segmentation. We also establish two benchmarks specific to the food domain: one for multi-turn dialogues that involve complex reasoning, and another for food-related reasoning segmentation. These benchmarks are crucial for assessing the reasoning and dialogue capabilities. Fine-tuned via reasoning instructions, FoodLMM shows outstanding performance in complex reasoning tasks and multi-turn dialogues about food-related subjects. 



%
%
\bibliographystyle{splncs04}
\bibliography{main}
\clearpage
\setcounter{page}{1}
\newcommand{\maketitlesupplementary}{
  \author{} 
  \institute{} 
  \maketitle 
  \vspace{-1.1cm} 
  \begin{center}
    Supplementary Material
  \end{center}
  \vspace{-0.8cm} 
}


\title{FoodLMM: A Versatile Food Assistant using Large Multi-modal Model} 
\maketitlesupplementary

\setcounter{page}{1}

\section{Details of Datasets}

\subsection{Stage 1: Public Food Datasets}

We utilize five public food datasets in the first training stage: 
\par
\noindent \textbf{VIREO Food-172~\cite{chen2016deep}} dataset contains around 100k food images from 172 categories and includes 353 ingredients. 
\par
\noindent \textbf{Recipe1M~\cite{salvador2017learning}} dataset includes about 922k recipes accompanied by 819k images, with some recipes corresponding to several images and others to none. This collection contains about 16k food ingredients.
\par
\noindent \textbf{Nutrition5k~\cite{thames2021nutrition5k}} offers visual data for about 5,000 dishes, including overhead RGB images and videos from four angles. Furthermore, this dataset also provides detailed nutritional information for each dish, covering total mass, total calories, and total macro-nutrient (fat, carbohydrate, protein) contents, along with specific data on the weight, calorie, and macro-nutrient content for every ingredient in each dish.
\par
\noindent \textbf{FoodSeg103~\cite{wu2021large}} dataset consists of 7,118 food images, each with ingredient-level fine-grained segmentation mask annotations. 
There are 103 ingredient types, and the number of ingredient mask annotations is 42,097.
\par
\noindent \textbf{UECFoodPixComplete~\cite{okamoto2021uec}} provides food category-level segmentation mask annotations for 10,000 images, containing a total of 102 food categories.

In section 3.1, we transform various public food datasets into the instruction-following format using our carefully designed instruction templates. The construction details for each task are introduced below.

\vspace{-0.5cm}
\begin{figure}[htbp]
  \centering
    \includegraphics[width=1\textwidth]{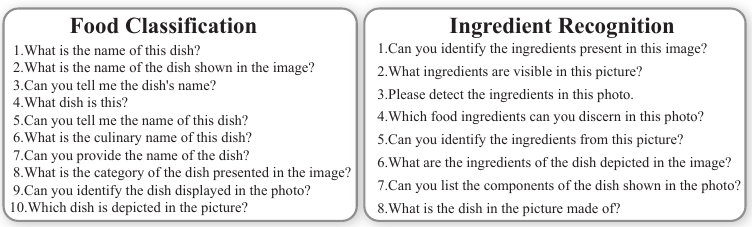} 
   \caption{Templates for Food Classification and Ingredient Recognition tasks.}
  \label{fig:fc_and_ir}
\vspace{-0.5cm}
\end{figure}

\noindent \textbf{Food Classification.}
We adopt the VIREO Food-172 to construct the instruction-following data for the food classification task. Our 10 question templates are detailed on the left of Figure~\ref{fig:fc_and_ir}.



\noindent \textbf{Ingredient Recognition.}
We use VIREO Food-172 and Recipe1M to develop instruction-following data for the ingredient recognition task. The templates are outlined on the right of Figure~\ref{fig:fc_and_ir}.



\begin{figure}[t!]
  \centering
    \includegraphics[width=1\textwidth]{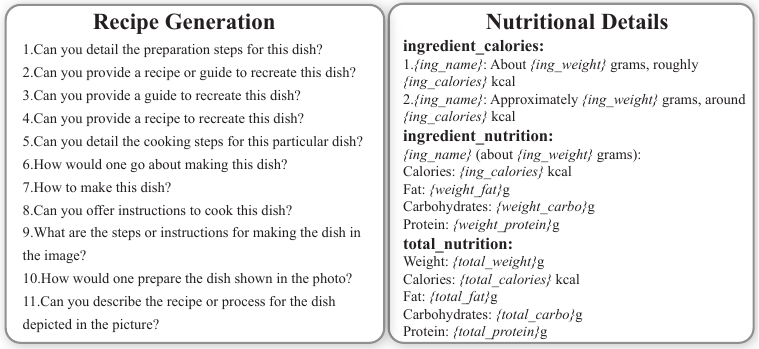}
   \caption{Templates for Recipe Generation task and outlining the nutritional details of food and its ingredients.}
  \label{fig:rg_and_NE_details}
\vspace{-0.5cm}
\end{figure}

\noindent \textbf{Recipe Generation.}
We construct instruction-following data for the recipe generation task using recipes from the Recipe1M dataset that have corresponding images. Moreover, consistent with~\cite{salvador2019inverse}, we map the 16k ingredients in Recipe1M to 1488. The employed query templates are provided on the left of Figure~\ref{fig:rg_and_NE_details}.

\begin{figure}[htbp]
\vspace{-0.3cm}
  \centering
    \includegraphics[width=1\textwidth]{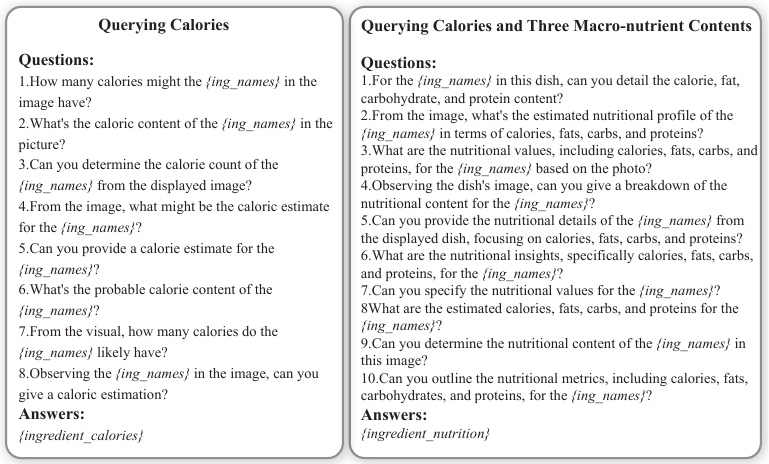}
    \vspace{-0.6cm}
   \caption{Templates for querying calories of any ingredient in the food and querying about calories and the three macro-nutrient contents of any ingredient in the food.}
  \label{fig:NE_2_and_4}
\vspace{-0.8cm}
\end{figure}

\noindent \textbf{Nutrition Estimation.}
To construct the nutrition estimation instruction-following data, we use the Nutrition5k dataset, which includes overhead RGB images and videos from various angles. 
Following the strategy described in~\cite{thames2021nutrition5k}, we sample every 5th frame from angle A and D videos for each dish. During training, for each dish, the probability of using the overhead RGB image is $70\%$ and the probability of randomly selecting a frame from the samples is $30\%$. We first establish three templates (on the right of Figure~\ref{fig:rg_and_NE_details}): 1) $ingredient\_calories$: the calorie content of ingredient (including weight and calories of the ingredient); 2) $ingredient\_nutrition$: the nutrition content of ingredient (including weight, calories, and three macro-nutrients); 3) $total\_nutrition$: the total nutrition content of the food (including the total weight, total calories, and the total content of the three macro-nutrients).

\begin{figure*}[htbp]
\vspace{-0.5cm}
  \centering
    \includegraphics[width=1\textwidth]{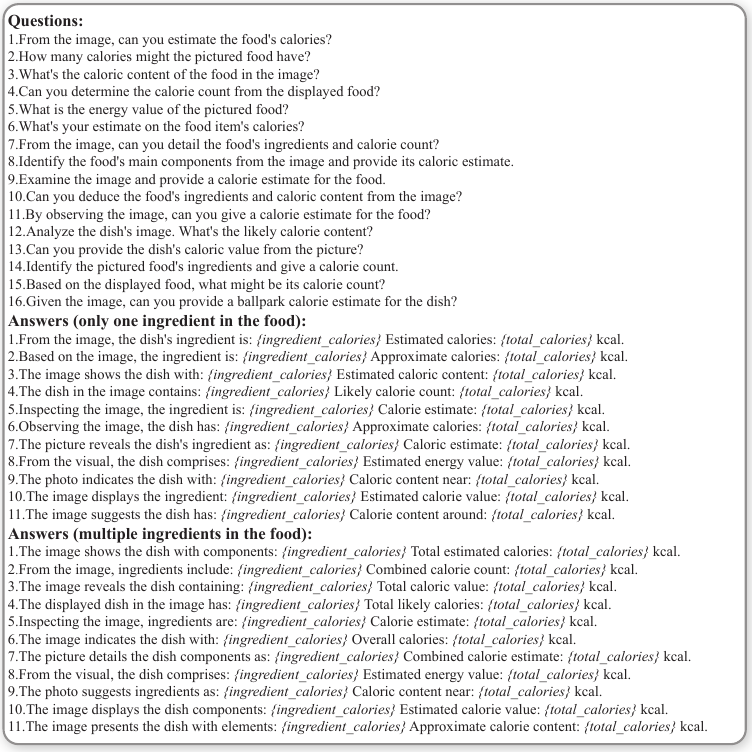}
   \caption{Templates for querying the total calories of the food and calories of each ingredient.}
  \label{fig:NE_1}
\vspace{-0.7cm}
\end{figure*}

\begin{figure}[htbp]
  \centering
    \includegraphics[width=1\textwidth]{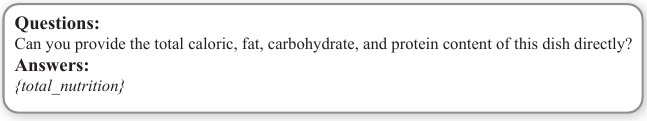}
   \caption{Templates for querying about the food's total calories and the content of the three macro-nutrients.}
  \label{fig:NE_5}
\vspace{-0.3cm}
\end{figure}

\begin{figure*}[t!]
  \centering
    \includegraphics[width=1\textwidth]{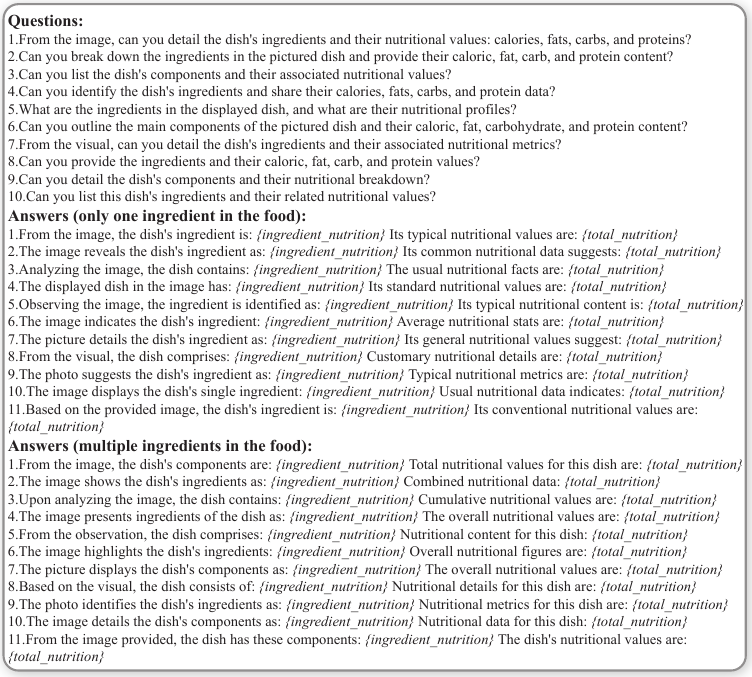}
   \caption{Templates for querying about the food's total calorie and macro-nutrient contents, as well as those of each individual ingredient.}
  \label{fig:NE_3}
 \vspace{-0.7cm}
\end{figure*}


\begin{figure}[t!]
  \centering
    \includegraphics[width=1\textwidth]{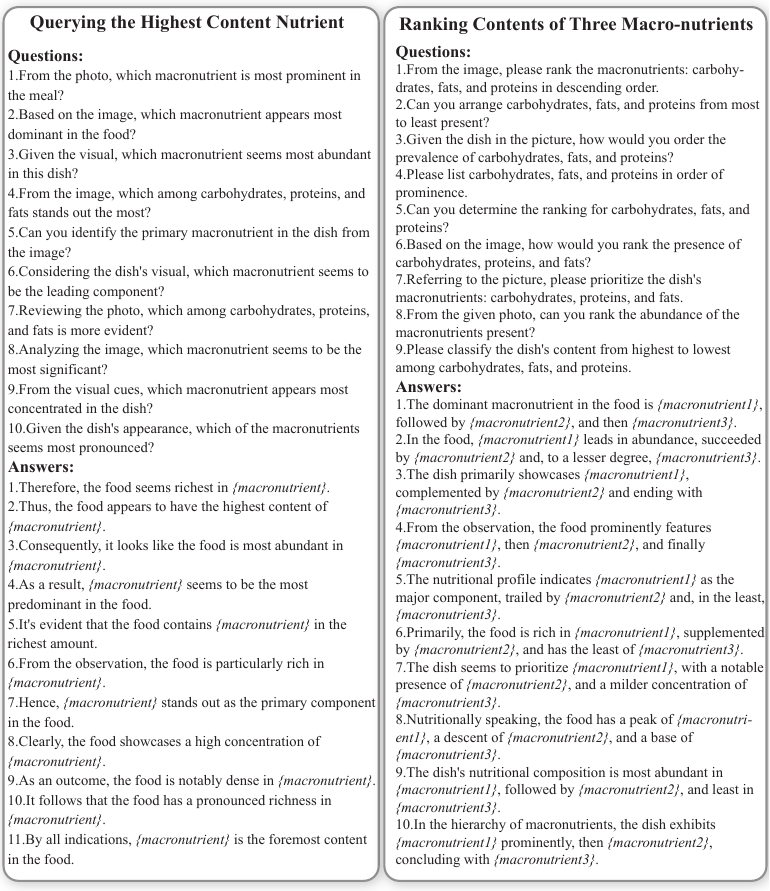}
   \caption{Templates for querying the nutrient with the highest content among the three macro-nutrients and ranking the contents of the three macro-nutrients in the food.}
  \label{fig:NE_6_and_7}
\vspace{-0.7cm}
\end{figure}

We design seven question-answer patterns for nutrition estimation instruction-following data. These are as follows: 
\par
1) Questions about the total calorie content of food and calories of each ingredient are shown in Figure~\ref{fig:NE_1}. $\{ingredient\_calories\}$ refers to the template in Figure~\ref{fig:rg_and_NE_details}, and $\{total\_calories\}$ indicates the total calorie content of the food. 
\par
2) Questions about calories of any ingredient in the food are presented on the left of Figure~\ref{fig:NE_2_and_4}, where $\{ingredient\_calories\}$ uses the template from Figure~\ref{fig:rg_and_NE_details} and $\{ing\_names\}$ represents one or several randomly chosen ingredients; 
\par
3) Questions about the total calorie content and three macro-nutrients in the food, along with each ingredient's calorie and macro-nutrient contents are outlined in Figure~\ref{fig:NE_3}. The $\{ingredient\_nutrition\}$ and $\{total\_nutrition\}$ correspond to the templates provided in Figure~\ref{fig:rg_and_NE_details}; 
\par
4) Questions about calories and three macro-nutrient contents of any ingredient in the food are displayed on the right of Figure~\ref{fig:NE_2_and_4}. $\{ingredient\_nutrition\}$ refers to the template in Figure~\ref{fig:rg_and_NE_details} and $\{ing\_names\}$ represents one or more randomly selected ingredients; 
\par
5) Questions about the total calories and the contents of the three macro-nutrients in the food are presented in Figure~\ref{fig:NE_5}, where $\{total\_nutrition\}$ refers to the template in Figure~\ref{fig:rg_and_NE_details};
\par
6) Questions about the nutrient with the highest content among the three macro-nutrients in the food are shown on the left of Figure~\ref{fig:NE_6_and_7}, with $\{macronutrient\}$ indicating the nutrient with the highest content;
\par
7) Ranking the contents of the three macro-nutrients in the food. Templates are displayed on the right of Figure~\ref{fig:NE_6_and_7}, where $\{macronutrient1\}$, $\{macronutrient2\}$, and $\{macronutrient3 \}$ represent the nutrients with the highest, second-highest, and third-highest content respectively.

\begin{figure}[htbp]
  \centering
    \includegraphics[width=1\textwidth]{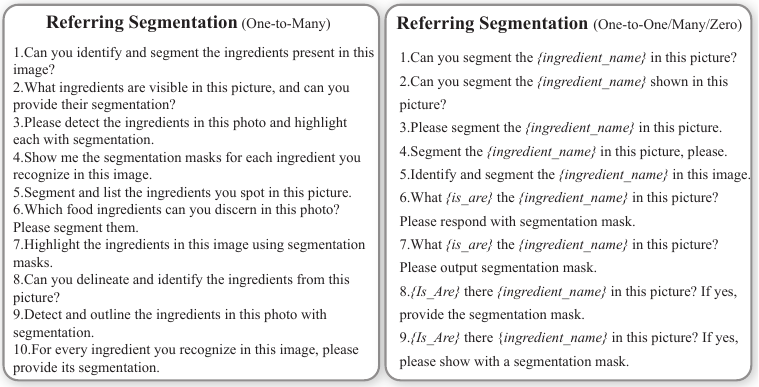}
   \caption{Templates for Referring Segmentation task.}
  \label{fig:tem_rs}
\vspace{-0.3cm}
\end{figure}
\noindent\textbf{Referring Segmentation.}
To train our FoodLMM to generate masks for multiple objects in query texts and identify absent items in images, similar to~\cite{liu2023gres}, we establish data containing multi-target (One-to-Many), no-target (One-to-Zero), and single-target (One-to-One) expressions. We divide the query templates into two groups to accommodate these three scenarios: 1) We employ the templates listed on the left of Figure~\ref{fig:tem_rs} to create instruction-following data for the One-to-Many scenario, where all templates guide the model to identify and segment all visible ingredients in food images. 2) Templates on the right of Figure~\ref{fig:tem_rs} are applicable to all three scenarios. For no-target data construction, $\{ingredient\_name\}$ represents one or multiple items absent in the image, and for multi-target and single-target data, it denotes one or several ingredients randomly selected from the image. $\{Is\_Are\}$ and $\{is\_are\}$ in the query text is automatically determined by the number of objects, to correctly use 'is' or 'are'.

\subsection{Stage 2: GPT-4 Generated Conversation Datasets}

\noindent \textbf{FoodDialogues.}
We utilize the Nutrition5k dataset to build our multi-turn conversation dataset FoodDialogues. Nutrition5k contains two types of visual information: RGB images from an overhead perspective and videos from multiple angles.
We select samples from Nutrition5k that contain both overhead perspective RGB images and good perspective videos (angle A or D) to construct our FoodDialogues. Specifically, each sample contains an overhead RGB image and a frame image extracted from the video as visual modality data. The data split of train and test sets is consistent with Nutrition5k.

We select seven representative food-related topics to generate multiple rounds of conversations, which are: 1. Nutrition and Ingredients, 2. Health and Diseases, 3. Calorie Calculation, 4. Metabolism, 5. Dietary Preferences and Allergies, 6. Dietary Planning and 7. Food Pairing and Substitution.

The number of question-answer (QA) pairs in a multi-round conversation sample is set between 3 to 5 rounds.
Taking into account the varying richness of ingredients in food images, we make the samples with more ingredients to cover more topics and have more rounds of QA, as detailed in Table~\ref{tab:fooddialogues_statistic}.

\begin{table}[t]
\centering
\scalebox{0.8}{
\begin{tabular}{lcccc}
\toprule
\textbf{Split} & \textbf{Ingredients} & \textbf{Topics} & \textbf{Rounds}  & \textbf{Samples} \\
\midrule
\multirow{3}{*}{Train} & 1$\sim$3 & 1 & 3  & 1,480\\
 & 4$\sim$10 & 3 & 4  & 4,130\\
 & 11$\sim$19 & 7 & 5 & 818 \\
\midrule
\multirow{3}{*}{Test} & 1$\sim$3 & 1 & 3  & 298\\
 & 4$\sim$10 & 3 & 4  & 647\\
 & 11$\sim$17 & 7 & 5 & 149 \\
\bottomrule
\end{tabular}
}
\vspace{0.27cm}
\caption{Dataset statistics of FoodDialogues.}
\label{tab:fooddialogues_statistic}
\vspace{-0.6cm}
\end{table}

\begin{figure}[h!]
\vspace{-0.5cm}
  \centering
\includegraphics[width=1\textwidth]{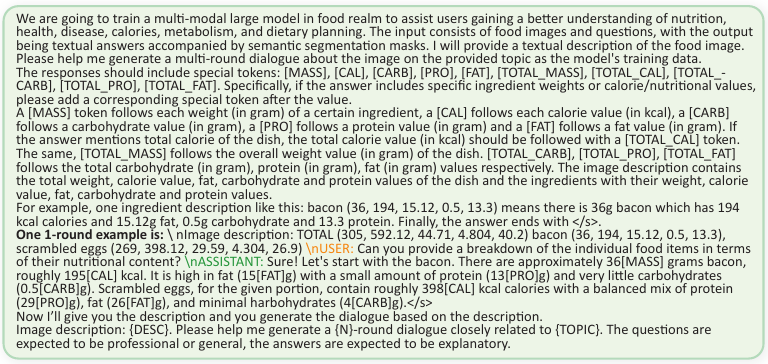} 
  \caption{Prompt template for generating FoodDlialogues.}
  \label{fig:fooddialogues_prompt}
\vspace{-0.5cm}
\end{figure}

\begin{figure*}[h!]
\vspace{-0.5cm}
  \centering
\includegraphics[width=1.0\textwidth]{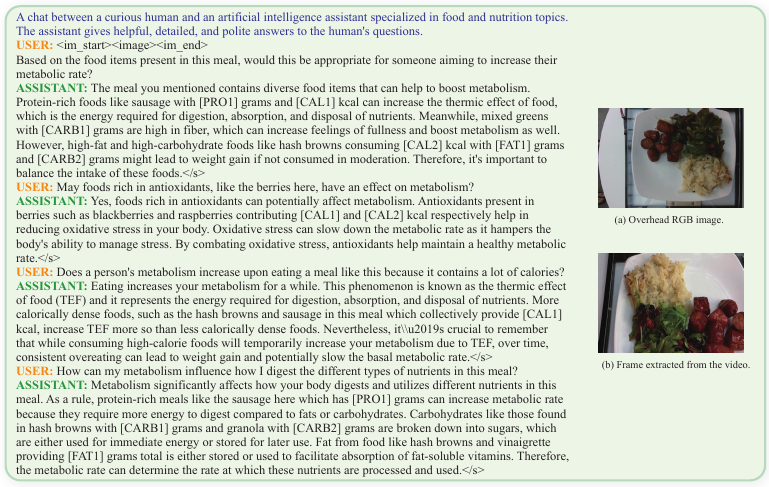} 
  \caption{An example from FoodDialogues.}
  \label{fig:fooddialogues_example1}
\vspace{-0.5cm}
\end{figure*}

We send the detailed nutritional information of the food images (i.e., the total mass, calories, carbohydrates, fat, and protein of the food and these values of each ingredient) to GPT-4 and instruct GPT-4 to generate multi-round dialogues of specific topics. 
The prompt template used for calling the GPT-4 API is provided in Figure~\ref{fig:fooddialogues_prompt}.
In the prompt, we instruct GPT-4 to mark each nutritional element value with our Nutrition Estimation task-specific tokens and give an example to ensure that the dialogue generated by GPT-4 conforms to the format we need.
We replace $\{DESC\}$, $\{TOPIC\}$ and $\{N\}$ in the template with the descriptions of different food images, the conversation topics that need to be generated, and the number of conversation turns respectively. 
For training FoodLMM, we take out the nutritional element values marked by task-specific tokens from the generated dialogue as supervision labels for the regression heads, while only task-specific tokens are retained in the text for training the LMM base. An example is shown in Figure~\ref{fig:fooddialogues_example1}.

\begin{figure}[htbp]
\vspace{-0.5cm}
  \centering
\includegraphics[width=1\textwidth]{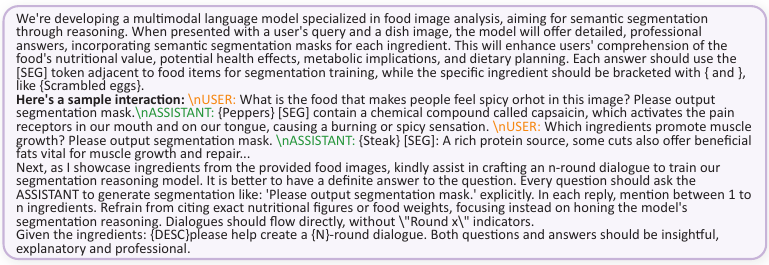} 
  \caption{Prompt template for generating FoodReasonSeg.}
  \label{fig:foodreasonseg_prompt}
\vspace{-0.5cm}
\end{figure}

\begin{figure*}[htbp]
\vspace{-0.5cm}
  \centering
\includegraphics[width=1.0\textwidth]{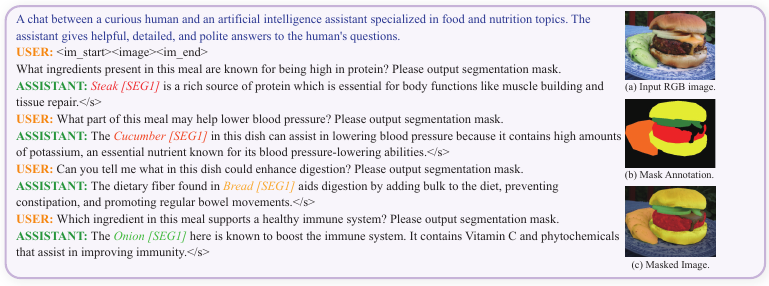} 
  \caption{An example from FoodReasonSeg.}
  \label{fig:foodreasonseg_example2}
\vspace{-0.7cm}
\end{figure*}

\noindent \textbf{FoodReasonSeg.}
We choose FoodSeg103, which contains fine-grained mask annotations for each ingredient in food images, to construct our food reasoning segmentation dataset FoodReasonSeg. 
Similar to FoodDialogues, FoodReasonSeg adopts a multi-turn conversation format. The more ingredients the food image contains, the greater the number of dialogue rounds.
To generate QA pairs with complex reasoning, we filter out food images with less than three ingredients. 
The statistics are shown in Table~\ref{tab:foodreasonseg_statistic}.
\begin{table}[htbp]
\vspace{-0.3cm}
\centering
\scalebox{0.8}{
\begin{tabular}{ccccc}
\toprule
\textbf{Ingredients} & \textbf{Rounds}  & \textbf{Train Samples}  & \textbf{Test Samples} \\
\midrule
 3 & 3  & 1,535  & 664\\
 4 & 4  & 1,224  & 450\\
 5 & 4  & 646  & 323\\
 6 & 4  & 304  & 108\\
 7 & 4  & 140 & 52\\
 8 & 4  & 39  & 18\\
 9 & 4  & 12   & 4\\
 10 & 4  & 6  & 1\\
 11 & 5  & 1  & 0\\
\bottomrule
\end{tabular}
}
\vspace{0.27cm}
\caption{Dataset statistics of FoodReasonSeg.}
\label{tab:foodreasonseg_statistic}
\vspace{-0.5cm}
\end{table}
We send the ingredient lists of the selected images to GPT-4, requesting it to generate multi-round dialogues consisting of QA pairs that require complex reasoning.
The prompt template is provided in Figure~\ref{fig:foodreasonseg_prompt}.
We ask GPT-4 to mark the ingredients mentioned in the answer with ${}$ so that we can retrieve the mask labels from the original dataset.
We also ask GPT-4 to generate segmentation tokens ([SEG]) in the answers for training FoodLMM.
During training, we remove the braces while retaining the name of ingredient and segmentation token.
An example is shown in Figure~\ref{fig:foodreasonseg_example2}.

\begin{figure*}[htbp]
\vspace{-0.5cm}
  \centering
\includegraphics[width=1.0\textwidth]{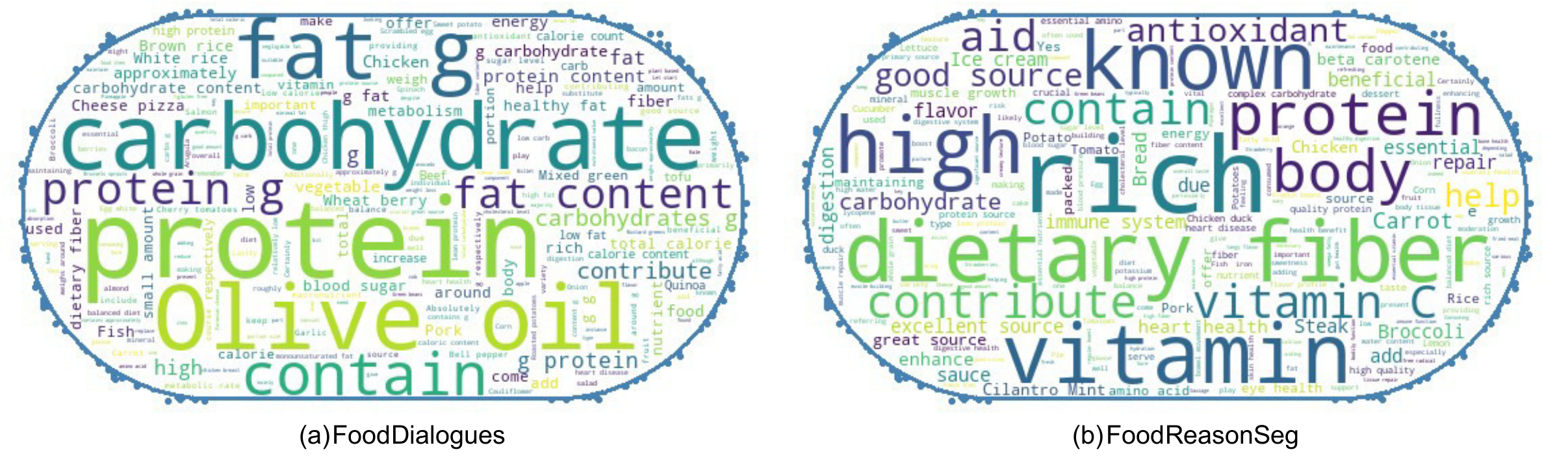} 
  \caption{Word cloud of FoodDialogues and FoodReasonSeg datasets.}
  \label{fig:wordcloud}
\vspace{-0.5cm}
\end{figure*}

Figure~\ref{fig:wordcloud} showcases the word cloud visualizations for FoodDialogues and FoodReasonSeg. In these visual representations, the prominence of each word is directly proportional to its frequency within the respective datasets.

\section{Experiment}
\subsection{Implementation Details}
\label{sec:implementation}

\noindent\textbf{Network Architecture.} Our model structure extends from LISA-7B~\cite{lai2023lisa}, which uses LLaVA-7B-v1-1~\cite{liu2023visual} as the LMM base and adopts ViT-H SAM~\cite{kirillov2023segment} as the segmentation encoder/decoder. The projection layer (MLP) for task-specific token embeddings is set to [256, 256, 256]. 
The corresponding channels of nutrition regression heads are set to [256, 1]. There are a total of 10 nutrition regression heads, which are used to regress the mass, calories, carbohydrates, fat, and protein for the dish and these values of each ingredient.


\noindent\textbf{Task-specific Tokens (Nutrition Estimation).}
We set 10 task-specific tokens corresponding to ten nutritional regression heads, 
We divide these 10 tokens into two categories: ingredient-level and dish-level.
The ingredient-level tokens include: $\langle \text{mass}\rangle$, $\langle \text{cal}\rangle$, $\langle \text{carb}\rangle$, $\langle \text{fat}\rangle$, $\langle \text{pro}\rangle$ for regressing the mass, calories, carbohydrates, fats, and proteins of any ingredient respectively.
And the dish-level tokens include: $\langle \text{total\_mass}\rangle$, $\langle \text{total\_cal}\rangle$, $\langle \text{total\_carb}\rangle$, $\langle \text{total\_fat}\rangle$, $\langle \text{total\_pro}\rangle$ for predicting the overall nutritional element values of the input food image. 
Note that in the implementation, we assigned numbers to the ingredient-level tokens, such as 
$\langle \text{mass}_1\rangle, \langle \text{mass}_2\rangle,  \ldots , \langle \text{mass}_n\rangle$ to differentiate the nutritional elements of different ingredients mentioned in the answers.
The LMM base (LLaVA) generates only these tokens without any specific value in response. 
Replace the corresponding tokens with the nutritional values obtained by regression heads to get the output answer.

\noindent\textbf{Task-specific Tokens (Referring/Reasoning Segmentation).}
For referring and reasoning segmentation tasks, we set task-specific token $\langle \text{seg}\rangle$. 
The LMM base (LLaVA) learns to assign $\langle \text{seg}\rangle$ tokens to the ingredients mentioned in the answer, and maps its corresponding hidden state to hidden embedding through the MLP. Hidden embedding serves as the segmentation prompt. 
The decoder is trained to generate a segmentation mask according to a $\langle \text{seg}\rangle$ hidden embedding.
Consistent with ingredient-level nutrition estimation tokens, $\langle \text{seg}\rangle$  tokens also use numerical numbers to distinguish different ingredients, i.e., $\langle \text{seg}_1\rangle, \langle \text{seg}_2\rangle,  \ldots , \langle \text{seg}_n\rangle$.
Summarize all generated masks to obtain the final segmented image containing multiple masks.

\begin{table}[htbp]
\centering
\scalebox{0.8}{
\begin{tabular}{l|l|l}
\toprule
\textbf{Model} & \textbf{Training} & \textbf{LoRA} \\
\midrule
 $\lambda_{\text{txt}}$=1.0 & precision=bf16 & r=8 \\
$\lambda_{\text{nutrition}}$=0.1 & optimizer=AdamW & alpha=16 \\
$\lambda_{\text{mask}}$=1.0 & learning\_rate=3e-4 & dropouts=5e-2 \\
$\lambda_{\text{MSE}}$=1.0 & weight\_decay=0.0 & \\
$\lambda_{\text {MAE}}$=1e-3 & warmup\_type=linear & \\
$\lambda_{\text {bce}}$=2.0 & warmup\_num\_steps=100 & \\
$\lambda_{\text {dice}}$=0.5 & gradient\_clipping=1.0 &\\
\bottomrule
\end{tabular}
}
\vspace{0.27cm}
\caption{Training hyperparameters.}
\label{tab:settings}
\end{table}

\noindent\textbf{Training details of Stage 1.}
We use LISA-7B-v1-explanatory model~\cite{lai2023lisa} as the initial checkpoint.
The training hyperparameter settings are detailed in Table~\ref{tab:settings}.
The sampling ratios of different datasets are shown in Table~\ref{tab:rate} (Stage 1).
Training takes 2 days on 4 NVIDIA A100 (40G) GPUs with batch size 4 and 10000 training steps.

\begin{table}[htbp]
\vspace{-0.3cm}
\centering
\scalebox{0.8}{
\begin{tabular}{l|c|c}
\toprule
\textbf{Dataset} & \textbf{Stage 1} & \textbf{Stage 2} \\
\midrule
VIREO Food-172~\cite{chen2016deep} & 5 & 5\\
Recipe1M~\cite{salvador2017learning} & 15 & 10\\
Nutrition5k~\cite{thames2021nutrition5k} & 10 & 10\\
FoodSeg103~\cite{wu2021large} & 6 & 6\\
UECFoodPixComplete~\cite{okamoto2021uec} & 4 & 4\\
FoodDialogues & —— & 45\\
FoodReasonSeg & —— &  30\\
\bottomrule
\end{tabular}
}
\vspace{0.27cm}
\caption{Sampling ratio of different datasets.}
\vspace{-0.5cm}
\label{tab:rate}
\end{table}

\noindent\textbf{Training details of Stage 2.}
Based on the Stage 1 model, we add the constructed FoodDialogues and FoodReasonSeg data sets to train Stage 2.
The training hyperparameters are consistent with Stage 1 (Table~\ref{tab:settings}), and the sampling ratio is shown in Table~\ref{tab:rate} (Stage 2).
Training takes 28 hours on 4 NVIDIA A100 (40G) GPUs with batch size 2 and 8000 training steps.

\noindent\textbf{Training/Fine-tuning details of different tasks.}
All of the fine-tuning experiments are conducted on 4 NVIDIA A100 (40G) GPUs.
For \textbf{Food Classification}, the fine-tuning step on Food-101 is set to 2500, taking 13 hours on average with batch size 4. 
For \textbf{Ingredient Recognition}, the fine-tuning step on VIREO Food-172 is set to 2500, taking 9 hours with batch size 4.
For \textbf{Recipe Generation}, the fine-tuning step on Recipe1M is set to 5000, taking 1 day with batch size 4.
For \textbf{Nutrition Estimation}, the fine-tuning step on Nutrition5k is set to 2500, taking 10 hours with batch size 2. Different types of question-answer pairs in the instruction templates are selected randomly both in the training and fine-tuning. For referring nutrition estimation questions (in Figure~\ref{fig:NE_2_and_4}), we randomly select any number of ingredients from the image for the query, with the number of ingredients in each query capped at 20.
For \textbf{Referring Segmentation}, FoodSeg103 and UECFoodPIXComplete are used together for fine-tuning. The step is set to 2500, taking 9 hours with batch size 2. 
The sampling rate of One-to-One/Many and One-to-Zero samples are set to 98\% and 2\% respectively both in the training and fine-tuning.
For the FoodSeg103 dataset, the probabilities of sampling questions from templates in Figures~\ref{fig:tem_rs} and Figure~\ref{fig:tem_rs} are 0.4 and 0.6, respectively.
For the UECFoodPixComplete dataset, since its annotations are not at ingredient-level and may not completely cover all ingredients, all the questions are sampled from templates in Figure~\ref{fig:tem_rs}.
For one-to-many referring queries, we randomly select several ingredients from the image for the query, with the number of ingredients in each query capped at 20.
For One-to-Zero referring queries, the maximum number of non-existent objects in a query is set to 3. The ratio of 1, 2 and 3 non-existent objects ($refer@1$, $refer@2$ and $refer@3$) in a One-to-Zero referring query is set to 2:1:1.
To increase the diversity of the queries, non-existent referring is hybridized with existent referring, i.e., there may be both present ingredients and absent objects in the same query.

\subsection{Additional results of the referring segmentation task}

\noindent\textbf{One-to-One / One-to-Many.}
We additionally report the referring segmentation performance of $refer@2$ and refer@4 situations in Table~\ref{tab:refer_results}. 
For the Complete Intersection over Union (cIoU) and the probability of the model correctly responding (ACC), as the number of ingredients in a query increases, both cIoU and ACC tend to decrease. This indicates that One-to-Many referring segmentation is more challenging than standard referring segmentation (One-to-One), and the difficulty escalates with an increasing number of objects referred to in the query.
The performance of our fine-tuned FoodLMM in the standard scenario ($refer@1$) far exceeds the current most powerful general segmentation large model LISA, improving 23\% on FoodSeg103 and 25\% on UECFoodPIXComplete, showing the strong professionalism of FoodLMM in the food realm. 
Even in the highly challenging $refer@4$ scenario, our FoodLMM maintains comparable performance (0.67 / 0.69) to the results of LISA in simple standard scenarios (0.65 / 0.72).
In contrast, LISA can only produce a single binary mask for all queries, unable to handle One-to-Many situations, thus limiting its applicability.

\begin{table*}[h!]
\vspace{-0.5cm}
\centering
\scalebox{0.73}{
\begin{tabular}{cc|cc|cc|cc|cc|cc}
\toprule
& & \multicolumn{2}{c|}{refer@1} & \multicolumn{2}{c|}{refer@2} & \multicolumn{2}{c|}{refer@3}  & \multicolumn{2}{c|}{refer@4} & \multicolumn{2}{c}{refer@5}\\
\multirow{-2}{*}{Dataset} & \multirow{-2}{*}{Method}  & cIoU  & Acc. & cIoU  & Acc. & cIoU & Acc. & cIoU  & Acc. & cIoU  & Acc. \\
\midrule
& LISA-7B~\cite{lai2023lisa} & 0.65 & \textbf{100.00}\% & —— & —— & —— & —— & —— & —— & —— & —— \\
& FoodLMM S1 & 0.78 & 99.16\% & 0.77 & 99.25\% & 0.73 & 98.99\%  & 0.66 & 99.05\% & 0.58 & 98.34\%  \\
\multirow{-3}{*}{FoodSeg103~\cite{wu2021large}} & FoodLMM FT & \textbf{0.80} & 98.69\%  & \textbf{0.78}  & \textbf{99.43}\% &\textbf{0.74}  & \textbf{99.51}\% & \textbf{0.67} & \textbf{99.48}\% & \textbf{0.62} & \textbf{99.29}\% \\
\midrule
& LISA-7B~\cite{lai2023lisa} & 0.72 & \textbf{100.00}\%  & —— & —— & —— & —— & —— & —— & —— & —— \\
& FoodLMM S1 & \textbf{0.90} & 99.70\%  & 0.74  & 98.02\% & 0.72  & 97.51\% & \textbf{0.71} & \textbf{98.15}\% & 0.64 & 98.10\% \\
\multirow{-3}{*}{UECFoodPIXComplete~\cite{okamoto2021uec}} & FoodLMM FT & \textbf{0.90} & 99.60\% & \textbf{0.76} & \textbf{98.42}\% & \textbf{0.76} & \textbf{99.38}\% & 0.69 & 96.30\% & \textbf{0.68} & \textbf{99.05}\% \\
\bottomrule
\end{tabular}
}
\vspace{0.27cm}
\caption{One-to-One/Many Referring Segmentation performance on FoodSeg103 and UECFoodPIXComplete.} 
\label{tab:refer_results}
\vspace{-0.7cm}
\end{table*}


\noindent\textbf{One-to-Zero.} 
We provide more results of One-to-Zero referring segmentation in Table~\ref{tab:refer_results_ood}.
LISA can not cope with this scenario. 
When querying for objects that do not exist in the image, LISA will always provide a segmentation mask, resulting in a zero accuracy rate for correctly responding (i.e., rejecting to segment non-existent items). 
In contrast, after being trained on Stage 1, our FoodLMM has been able to correctly reject most non-existent objects in the queries (73.07\% to 58.80\% on FoodSeg103 and 77.50\% to 52.30\% on UECFoodPIXComplete from refer@1 to refer@3), though its accuracy declines as the number of non-existent items increases.
However, after fine-tuning, the accuracy of correctly responding remains unaffected even with an increase in the number of non-existent items queried.

\begin{table}[!t]
\centering
\scalebox{0.73}{
\begin{tabular}{cc|c|c|c}
\toprule
Dataset & Method & $refer@1$ & $refer@2$ & refer@3  \\
\midrule
& LISA-7B~\cite{lai2023lisa} & 0.00\% & 0.00\% & 0.00\%\\
& FoodLMM S1 & 73.07\% & 59.91\% & 58.80\% \\
\multirow{-3}{*}{FoodSeg103} & FoodLMM FT & \textbf{87.87}\% & \textbf{89.88}\% & \textbf{88.99}\% \\
\midrule
& LISA-7B~\cite{lai2023lisa} & 0.00\% & 0.00\% & 0.00\%\\
& FoodLMM S1 & 77.50\% & 57.00\% & 52.30\% \\
\multirow{-3}{*}{UECFoodPIXComplete} & FoodLMM FT & \textbf{83.7}\% & \textbf{85.8}\% & \textbf{83.5}\% \\
\bottomrule
\end{tabular}
}
\vspace{0.27cm}
\caption{One-to-Zero Referring Segmentation performance on FoodSeg103 and UECFoodPIXComplete.} 
\label{tab:refer_results_ood}
\vspace{-0.3cm}
\end{table}

\begin{table*}[htbp]
\centering
\scalebox{0.62}{
\begin{tabular}{cc|c|c|c|cc|cc|cc|cc|cc}
\toprule
& & \multicolumn{3}{c|}{One-to-Zero} & \multicolumn{10}{c}{One-to-One/One-to-Many} \\
& & $refer@1$ & $refer@2$ & refer@3 & \multicolumn{2}{c|}{$refer@1$} & \multicolumn{2}{c|}{$refer@2$} & \multicolumn{2}{c|}{refer@3}  & \multicolumn{2}{c|}{refer@4} & \multicolumn{2}{c}{refer@5} \\
\multirow{-3}{*}{Dataset} & \multirow{-3}{*}{Method} & Acc. & Acc. & Acc.  & cIoU & Acc. & cIoU & Acc. & cIoU & Acc. & cIoU & Acc. & cIoU & Acc.\\
\midrule
& ratio=2\% & 73.07\% & 59.91\% & 58.80\% & 0.78 & \textbf{99.16}\% & \textbf{0.77} & \textbf{99.25}\% & \textbf{0.73} & \textbf{98.99}\%  & \textbf{0.66} & \textbf{99.05}\% & \textbf{0.58} & \textbf{98.34}\% \\
\multirow{-2}{*}{FoodSeg103~\cite{wu2021large}} & ratio=20\% & \textbf{97.28}\% & \textbf{79.11} & \textbf{66.56}\% & \textbf{0.85} & 28.57\%  & 0.46  & 58.56\% &  0.48 & 67.53\% & 0.51 & 74.53\% & 0.48 & 75.73\%  \\
\midrule
& ratio=2\% & 77.50\% & 57.00\% & 52.30\% & \textbf{0.90} & \textbf{99.70}\% & \textbf{0.74} & \textbf{98.02}\% & \textbf{0.72} & \textbf{97.51}\% & \textbf{0.71} & \textbf{98.15}\% & \textbf{0.64} & \textbf{98.10}\%  \\
\multirow{-2}{*}{UECFoodPIXComplete~\cite{okamoto2021uec}} & ratio=20\% & \textbf{99.30}\% & \textbf{74.90} & \textbf{65.00}\% & 0.89 & 7.00\% & 0.54 & 52.00\% & 0.65 & 59.00\% & 0.59 & 61.11\% & 0.59 & 68.57\% \\
\bottomrule
\end{tabular}
}
\vspace{0.27cm}
\caption{Ablation study of One-to-Zero referring data ratio.} 
\label{tab:ablation}
\vspace{-0.6cm}
\end{table*}

\subsection{Ablation experiment of the ratio of one-to-zero samples}
The above results (Table~\ref{tab:refer_results} and Table~\ref{tab:refer_results_ood}) show that by adding only a minimal amount (2\%, detailed in Section~\ref{sec:implementation}) of non-existent referring (One-to-Zero) queries in the training, our model exhibits excellent correct response accuracy  (ACC) for One-to-Zero queries (Table~\ref{tab:refer_results_ood}), while the response accuracy of One-to-One/Many queries is only slightly affected (Table~\ref{tab:refer_results}).
The results in Table~\ref{tab:ablation} show that increasing the ratio of non-existent referring queries can significantly improve the response accuracy for One-to-Zero queries. But it catastrophically damages the model's ability to correctly respond and segment for One-to-One/One-to-Many queries.
The results in Table~\ref{tab:ablation} show that increasing the ratio of non-existent referring data can significantly improve the response accuracy for One-to-Zero queries. But it catastrophically damages the model's ability to correctly respond and segment for One-to-One/One-to-Many queries.
When conducting One-to-One queries, the response accuracy (ACC) of the ratio=20\% model on the two datasets are only 28.57\% and 7.00\%, which are 71\% and 99\% decreases compared to the ratio=2\% model (99.16\% and 99.70\%). 
Despite the cIoU score is still comparable to the 2\% model, this is because the model rejects the majority of queries, responding only to a few.
As the number of referred ingredients in One-to-Many queries increases, the response accuracy of the 20\% model starts to rise.
This is because queries with only one non-existent object are the most common in training samples, leading the model to be more inclined to refuse to provide masks for any $refer@1$ query.
Queries with multiple non-existent objects are less frequent in training samples, so the response accuracy for One-to-Many queries paradoxically increases, but still far less than the 2\% model. 
This suggests that incorporating a small number of non-existent referring queries during training is enough for the model to develop a considerable one-to-zero capability. Conversely, a higher proportion of non-existent referring queries biases the model towards refusing to respond, even for correct referring. 
Furthermore, this ablation experiment shows that the impact of imbalanced data distribution (long tail) on LMM still exists~\cite{kandpal2023large} and is worthy of further exploration.

\subsection{Additional Results of Food VQA tasks}


We present detailed results comparing our FoodLMM with other methods in food classification, ingredient recognition, and recipe generation tasks. 
\par
Food-101~\cite{bossard2014food} is acknowledged as a benchmark in computer vision, widely used in various domains. We conduct experiments to verify the classification capability of our model on this dataset. Table~\ref{tab:food_classification_results} provides detailed evaluation results which include performances of existing methods with their most effective backbone. As shown in Table~\ref{tab:food_classification_results}, FoodLMM outperforms existing methods in Top-1 classification accuracy, at 93.93\%.

\begin{table}[htbp]
\centering
\scalebox{0.8}{
\begin{tabular}{ll}
\toprule
\textbf{Method} & \textbf{Acc.} \\
\midrule
WARN(WRN-50)~\cite{rodriguez2019pay}  & 85.50  \\
PMG(ResNet50)~\cite{du2020fine} & 86.93  \\
WS-DAN(Inceptionv3)~\cite{hu2019see} & 88.90 \\
DCL(ResNet50)~\cite{chen2019destruction} & 88.90 \\
PAR-Net(ResNet101)~\cite{qiu2022mining} & 89.30\\
SGLANet(SENet154)~\cite{min2020isia} & 89.69 \\
Inception v4 @ 448px~\cite{kornblith2019better} & 90.00 \\
IG-CMAN(SENet154)~\cite{min2019ingredient} & 90.37 \\
MSMVFA(SENet154)~\cite{jiang2019multi} & 90.59 \\
PRENet (SENet154+Pretrained)~\cite{min2023large} & 91.13 \\
\midrule
FoodLMM FT   & 93.93 \\
\bottomrule
\end{tabular}
}
\vspace{0.27cm}
\caption{Performance comparison on Food-101~\cite{bossard2014food}.}
\label{tab:food_classification_results}
\vspace{-0.5cm}
\end{table}

\begin{table}[htbp]
\vspace{-0.5cm}
\centering
\scalebox{0.8}{
\begin{tabular}{ll}
\toprule
\textbf{Method} & \textbf{F1 score} \\
\midrule
DSDL(Resnet101)~\cite{zhou2021deep}    & 49.44    \\
AFN+BFL(Resnet101)~\cite{liu2020food} & 58.53    \\
ML-GCN(Resnet101)~\cite{chen2019multi}  & 55.32    \\
SGTN(Resnet101)~\cite{vu2020privacy}    & 53.97    \\
MGTN(Resnet101)~\cite{nguyen2021modular}    & 55.08    \\
FCI-MTL(Resnet101)~\cite{wang2022ingredient} & 59.56    \\
DB-Focal(Resnet50)~\cite{wu2020distribution} & 60.93    \\
ASL(Resnet101)~\cite{ridnik2021asymmetric}     & 63.08    \\
D-Mixup(Resnet101)~\cite{gao2022dynamic} & 63.16    \\
CACLNet(Resnet101)~\cite{luo2023ingredient} & 65.71    \\
\midrule
FoodLMM FT         & 68.97   \\
\bottomrule
\end{tabular}
}\vspace{0.27cm}
\caption{Performance comparison on VIREO Food-172~\cite{chen2016deep}.}
\label{tab:ingredients_recognition_results}
\vspace{-0.6cm}
\end{table}

\par
Table~\ref{tab:ingredients_recognition_results} showcases FoodLMM's proficiency in ingredient recognition. We compare our model against methods with strong performance in ingredient recognition, such as AFN+BFL~\cite{liu2020food}, FCI-MTL~\cite{wang2022ingredient}, D-Mixup~\cite{gao2022dynamic}, and CACLNet~\cite{luo2023ingredient}. Additionally, we evaluate the efficacy of general multi-label classification approaches like MGTN~\cite{nguyen2021modular}, ASL~\cite{ridnik2021asymmetric}, and DB-Focal~\cite{wu2020distribution} in ingredient recognition task. For each existing method, we chose the most suitable backbone. FoodLMM shows the highest performance in this task, surpassing the previous best CACLNet~\cite{luo2023ingredient}, CACLNet, by 4.96\%.

\begin{table}[htbp]
\centering
\scalebox{0.8}{
\begin{tabular}{ll}
\toprule
\textbf{Method} & \textbf{F1 score} \\
\midrule
DSDL(Resnet101)~\cite{zhou2021deep}    & 49.44    \\
AFN+BFL(Resnet101)~\cite{liu2020food} & 58.53    \\
ML-GCN(Resnet101)~\cite{chen2019multi}  & 55.32    \\
SGTN(Resnet101)~\cite{vu2020privacy}    & 53.97    \\
MGTN(Resnet101)~\cite{nguyen2021modular}    & 55.08    \\
FCI-MTL(Resnet101)~\cite{wang2022ingredient} & 59.56    \\
DB-Focal(Resnet50)~\cite{wu2020distribution} & 60.93    \\
ASL(Resnet101)~\cite{ridnik2021asymmetric}     & 63.08    \\
D-Mixup(Resnet101)~\cite{gao2022dynamic} & 63.16    \\
CACLNet(Resnet101)~\cite{luo2023ingredient} & 65.71    \\
\midrule
FoodLMM FT         & 68.97   \\
\bottomrule
\end{tabular}
}\vspace{0.27cm}
\caption{Performance comparison on VIREO Food-172~\cite{chen2016deep}.}
\label{tab:ingredients_recognition_results}
\end{table}

\par
Advancements in ingredient recognition techniques have enabled recipe generation not solely reliant on retrieval methods. For example, Chef Transformer~\cite{ChefTransformer} generates cooking instructions directly from ground-truth ingredients. FoodLMM's performance in recipe generation, shown in Table~\ref{tab:recipe_generation_results}, confirms the large language model's ability to generate recipes based on food images. Notably, FoodLMM creates recipes without relying on additional information such as ingredient lists or the food category. Our model achieves SacreBLEU and Rouge-L scores of 6.24 and 39.96, respectively, marking improvements of 3.65\% and 21.46\% over FIRE~\cite{chhikara2023fire}.
\begin{table}[htbp]
\centering
\scalebox{0.8}{
\begin{tabular}{lll}
\toprule
\textbf{Method}  & \textbf{SacreBLEU} & \textbf{ROUGE-L} \\
\midrule
Chef Transformer~\cite{ChefTransformer} & 4.61 & 17.54 \\
InverseCooking~\cite{salvador2019inverse}  & 5.48 & 19.47 \\
FIRE~\cite{chhikara2023fire} & 6.02 & 21.29 \\
\midrule
FoodLMM FT       & 6.24 & 36.96 \\
\bottomrule
\end{tabular}
}
\vspace{0.27cm}
\caption{Performance comparison on Recipe1M~\cite{salvador2017learning}.}
\label{tab:recipe_generation_results}
\end{table}


\subsection{More Qualitative Examples}
\begin{figure*}[htbp]
  \centering
\includegraphics[width=1.0\textwidth]{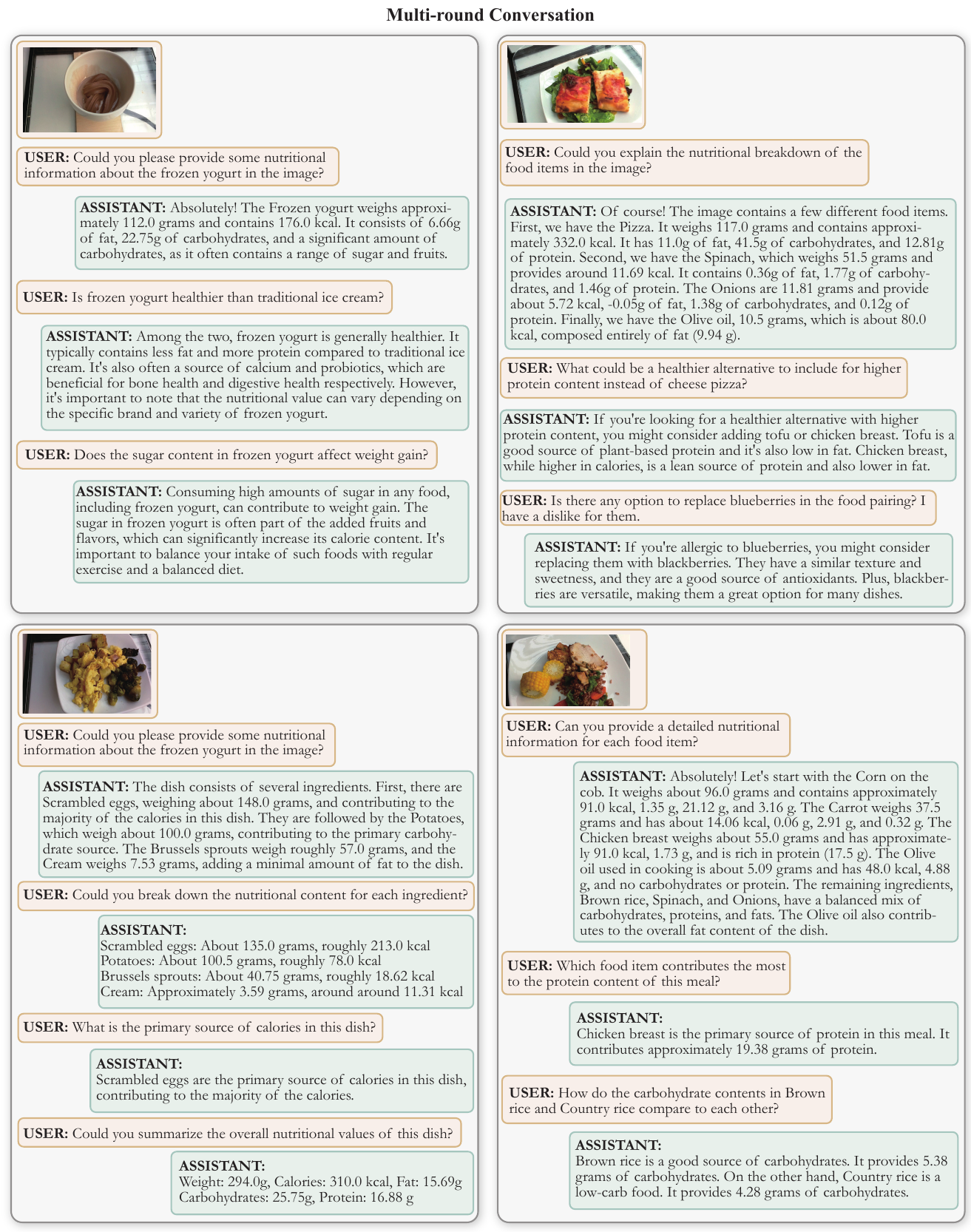} 
  \caption{Qualitative examples of multi-round conversation.}
  \label{fig:FoodDialogues_all}
\end{figure*}

\begin{figure*}[htbp]
  \centering
\includegraphics[width=1.0\textwidth]{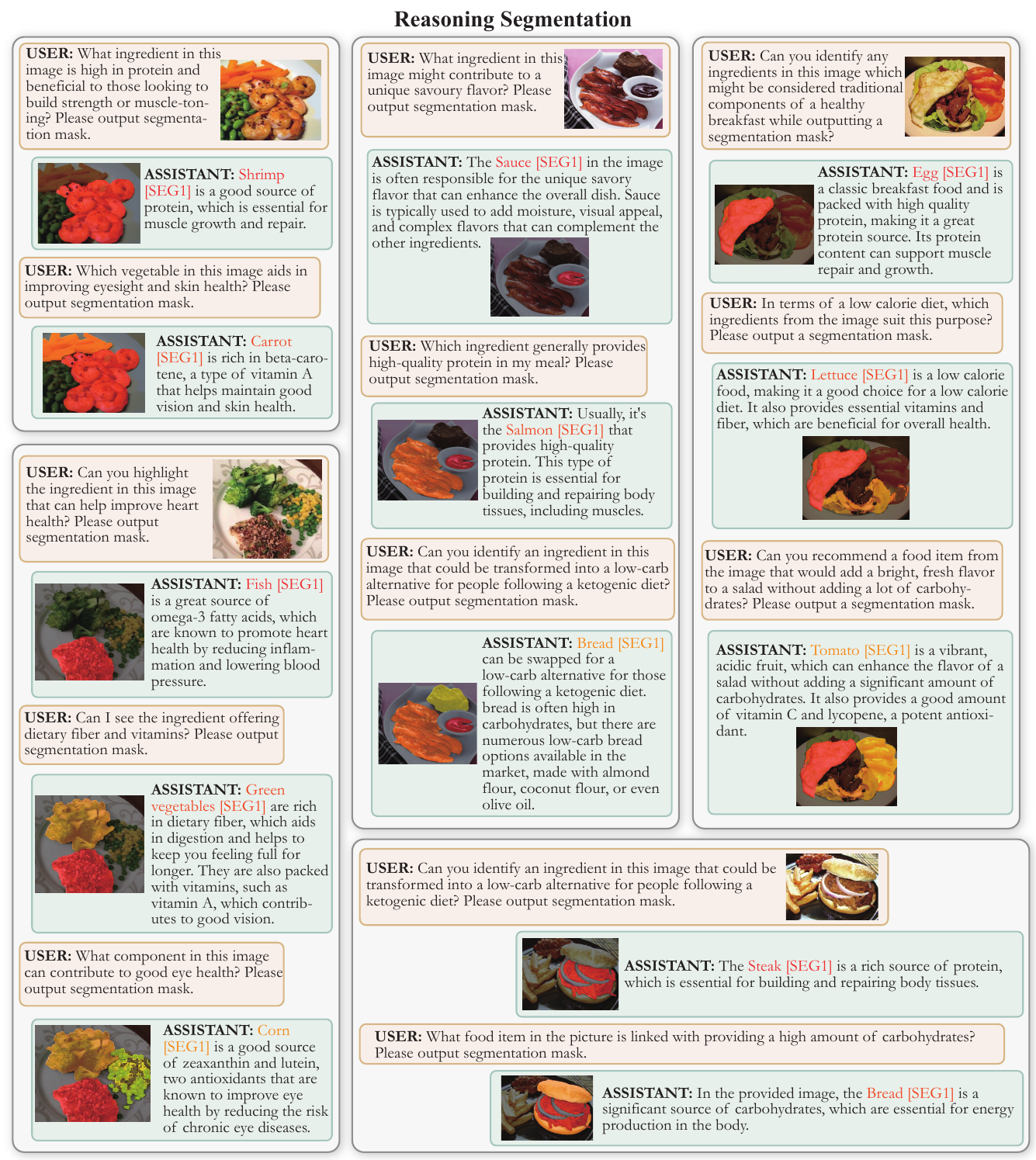} 
  \caption{Qualitative examples of reasoning segmentation.}
  \label{fig:tasks_FoodReasonSeg}
\end{figure*}

Figure~\ref{fig:FoodDialogues_all} present additional qualitative results from multi-turn dialogue tasks on nutrition-related topics. In Figure~\ref{fig:tasks_FoodReasonSeg}, we display additional qualitative examples from the reasoning segmentation task.

\end{document}